\PassOptionsToPackage{hyphens}{url}
\documentclass{article}
\usepackage{PRIMEarxiv}
\usepackage[utf8]{inputenc}
\usepackage[T1]{fontenc}
\usepackage{natbib}
\usepackage{xcolor}
\providecolor{primeblue}{RGB}{0,51,102}
\usepackage{hyperref}
\usepackage{url}
\usepackage{booktabs}
\usepackage{longtable}
\usepackage{multirow}
\usepackage{array}
\usepackage{float}
\usepackage{flafter}
\usepackage[section]{placeins}
\usepackage{amsmath}
\usepackage{amssymb}
\usepackage{amsfonts}
\usepackage[expansion=false]{microtype}
\usepackage{fancyhdr}
\usepackage{graphicx}
\usepackage{enumitem}
\usepackage{xspace}
\usepackage{subcaption}
\graphicspath{{figures/}}
\hypersetup{colorlinks=true,linkcolor=primeblue,citecolor=primeblue,urlcolor=primeblue}

\newcommand{\method}{ModernELECTRA\xspace}
\newcommand{\enmodel}{EdgeCoder-31M\xspace}
\newcommand{\mlmodel}{EdgeCoder-Multilingual-61M\xspace}

\title{ModernELECTRA: Replaced-Token Detection with Boundary-Window
Reconstruction for Compact Encoders that Classify and Embed}

\author{
\textbf{Mykhailo Shtopko$^{1}$, Ihor Stepanov$^{1}$, Dmytro Vodianytskyi$^{1}$, Oleksandr Lukashov$^{1}$}\\
$^{1}$Knowledgator Engineering, Kyiv, Ukraine \\
\textbf{Correspondence:} \texttt{mykhailoshtopko@knowledgator.com}
}

\begin{document}
\maketitle

\begin{abstract}
ELECTRA-style replaced-token detection (RTD) is a sample-efficient
objective for pre-training text encoders, and RTD-trained models such as
DeBERTaV3 are strong on classification benchmarks. The same models, however,
are known to be poor backbones for embedding tasks such as retrieval and
semantic similarity: the binary ``original vs.\ replaced'' signal does not ask
the final layer to retain the token identity and contextual content that
dense representations rely on. We introduce \method, a pre-training recipe
that keeps RTD as the main objective and adds a lightweight auxiliary task,
\emph{boundary-window reconstruction}. For every token, we sample a window of
3--12 tokens that strictly contains it and reconstruct the token's original
identity \emph{only} from the discriminator's hidden states at the two window
boundaries. The task costs one small FFN at training time and is discarded
afterwards, yet it forces every final-layer state to carry information about
its neighbourhood. Combined with a frozen, separately pre-trained generator,
a temperature curriculum for hard negatives and a 512$\to$8192 length
curriculum, we train two compact encoders: \enmodel, an 8-layer English
model, and \mlmodel, a 16-layer model covering 46 languages, each on about
1T tokens. They reach 86.0 and 85.1 on GLUE. The English model outperforms
Ettin-Enc-32m (83.5) and MiniLM-L12 (84.6) on GLUE while matching them on
MTEB English v2 (51.1), and the multilingual model outperforms the
2.3$\times$ larger mmBERT-small on GLUE and Multilingual MiniLM and
mDistilBERT on XTREME over the languages it covers. We also describe
FlashDeBERTa, our open-source fused implementation of DeBERTa's disentangled
attention that stores no $n\times n$ score matrix: at 4096 tokens it is
10$\times$ faster and uses 93\% less memory than the reference
implementation, and it loads existing DeBERTa checkpoints unchanged.
\end{abstract}

{\sloppy}

\section{Introduction}
\label{sec:intro}

Bidirectional encoders remain the workhorse of production NLP: they power
classification, named-entity recognition, re-ranking, and the
embedding models~\citep{reimers2019sbert} behind retrieval-augmented
generation~\citep{lewis2020rag}. They are also the backbones of lightweight
zero-shot models for information extraction and classification such as
GLiNER~\citep{zaratiana2024gliner,stepanov2024gliner} and
GLiClass~\citep{stepanov2025gliclass}, which typically build on DeBERTaV3. In
these settings the encoder is called on every document and every query, so
small models with long context are worth far more than their benchmark
position alone would suggest.

Two pre-training families dominate. Masked language modelling (MLM), used by
BERT~\citep{devlin2019bert}, RoBERTa~\citep{liu2019roberta}, the first
DeBERTa models~\citep{he2021deberta}, ModernBERT~\citep{warner2024modernbert}
and mmBERT~\citep{marone2025mmbert},
trains on the 15--30\% of positions that are masked~\citep{wettig2023mask}. Replaced-token detection
(RTD), introduced by ELECTRA~\citep{clark2020electra} and adopted by
DeBERTaV3~\citep{he2023debertav3}, trains a discriminator to decide for
\emph{every} token whether a small generator replaced it. RTD is markedly more
sample-efficient~\citep{clark2020electra}, and RTD models remain among the strongest encoders on
classification: DeBERTaV3-base reaches 88.1 GLUE, ahead of BERT-base (84.7)
and RoBERTa-base (86.4)~\citep{warner2024modernbert}.

\paragraph{The RTD embedding gap.}
It is well established that RTD encoders transfer poorly to embedding
tasks, and to retrieval in particular. In the controlled comparison of~\citet{warner2024modernbert},
in which every backbone is fine-tuned on MS\,MARCO with the same recipe as a
dense retriever that encodes each text into one embedding (DPR-style),
DeBERTaV3-base reaches only 20.2 nDCG@10 on BEIR, about
half of BERT-base (38.9) and ModernBERT-base (41.6). DeBERTaV3-large reaches
25.6, against 44.0 for ModernBERT-large. The authors explicitly note
DeBERTaV3's ``strong results \ldots\ on classification tasks but weak ones on
retrieval''. \citet{rep2024electra} make the same observation for semantic
textual similarity: ELECTRA's sentence embeddings are weak enough that the
community ``tacitly stopped'' using them, and the drop is concentrated in the
discriminator's \emph{last} layer, which underperforms earlier layers.
The gap also appears on MTEB. After the same contrastive fine-tuning for all
models, \citet{breton2025neobert} report an average of 26.9 for
DeBERTaV3-base and 27.1 for DeBERTaV3-large on MTEB English, against
45.0--49.1 for BERT, RoBERTa and ModernBERT of the same sizes.
\citet{antoun2025modernbert} compare French encoders pre-trained on the same
corpus: the RTD-trained CamemBERTaV2 averages 31.2 on MTEB-French, against
51.7 for the MLM-trained CamemBERTv2, although it is the best of the compared
models on classification, NER and question answering.

The cause is the objective itself. The generator proposes plausible
replacements, often tokens that are semantically close to the original, so
the discriminator has to tell near-synonyms apart in context. This is what
makes RTD encoders strong on classification. But RTD asks the final layer
for only one bit per token. Nothing in the loss requires the final hidden
state of a token to encode \emph{which} token it is or what its neighbours
are, so the discriminator can discard everything that it does not need to
detect a replacement. MLM, in contrast, must reconstruct the identity of
each masked token from its context, which keeps lexical and contextual
content in the final layer. Mean-pooled sentence embeddings and
token-level matching rely on exactly this content.

\citet{antoun2025modernbert} instead attribute the failure to the
architecture, and hypothesise that DeBERTa lacks the mechanisms to build a
document representation by pooling token states. The available evidence does
not support this. DeBERTa models pre-trained with MLM show no such gap when
they are fine-tuned as embedding models~\citep{weller2025ettin}, and the
models in this paper use the same disentangled attention and match MLM
encoders of their size on embedding tasks (\S\ref{sec:results}).

\paragraph{This work.}
We keep RTD as the main training signal and add a small auxiliary task that
restores the missing information: \emph{boundary-window reconstruction}
(\S\ref{sec:method:window}). For every token we sample a window of 3--12 tokens
that strictly contains it, and reconstruct the token's \emph{original}
identity only from the final hidden states at the two window
boundaries. Every token is reconstructed in this way, and every token also
serves as a boundary of the windows around its neighbours. Each final-layer
state is therefore trained to carry information about the tokens up to ten
positions to its left and right. The
reconstruction is decoded through the frozen generator's LM transform against
the discriminator's own token-embedding table, so the auxiliary task adds
only one FFN at training time and nothing at inference.

With this recipe, which we call \method, we train two compact encoders:
\enmodel, an 8-layer English model, and \mlmodel, a 16-layer model for 46
languages (\S\ref{sec:method}--\ref{sec:training}). Both are strong on
classification for their size and, unlike earlier RTD models, match MLM
encoders of the same size on English embedding tasks (\S\ref{sec:results}).

We focus on compact models on purpose. The quality of an encoder saturates
as it grows: in the Ettin suite, where all sizes are trained on the same
data, each doubling of the model adds about four GLUE points between 17M and
68M parameters, while going from 400M to 1B parameters adds
0.8~\citep{weller2025ettin}. Small models are therefore the range where
quality is furthest from saturation and where a better pre-training recipe
has the largest effect. They are also the models that run on CPUs and edge
devices, and interest in this range is growing: recent encoder families
ship small variants~\citep{weller2025ettin,marone2025mmbert}, and recent embedding
models are designed for on-device use~\citep{vera2025embeddinggemma}. For
RTD encoders this range also has a specific gap: to our knowledge, there
was no compact model that keeps the classification quality of RTD and can
also be used for embeddings.

Training a DeBERTa model on 8192-token inputs also requires an efficient
implementation of its attention, because the reference implementation stores
several $n\times n$ matrices per head and cannot use standard fused kernels.
We therefore describe FlashDeBERTa (\S\ref{sec:flash}), our open-source set of fused GPU
kernels for disentangled attention with support for batches without padding.
It is several times faster and far more memory-efficient on long inputs,
and it works with existing DeBERTa-v2 and DeBERTa-v3 checkpoints.

\section{Related Work}
\label{sec:related}

\paragraph{Encoder pre-training objectives.}
BERT~\citep{devlin2019bert} introduced masked language modelling, and
RoBERTa~\citep{liu2019roberta} showed that more data, larger batches and
dynamic masking improve it substantially.
ELECTRA~\citep{clark2020electra} replaced MLM with replaced-token detection
against a jointly trained small generator and reported large gains in sample
efficiency, especially for small models.
DeBERTa~\citep{he2021deberta} introduced disentangled content/position
attention, and DeBERTaV3~\citep{he2023debertav3} combined it with RTD and
gradient-disentangled embedding sharing.
Span-oriented objectives are closely related to our auxiliary task.
SpanBERT~\citep{joshi2020spanbert} predicts every token of a masked span from
the representations of the span's two boundary tokens and a position
embedding (the span-boundary objective, SBO). Boundary-window reconstruction
borrows this boundary-only bottleneck but differs in three ways: it runs on
the \emph{discriminator's} representations of a corrupted, unmasked
sequence; it samples a window for \emph{every} token rather than for masked
spans; and it uses linear interpolation of the two boundary states by
relative offset instead of a learned position embedding.

\paragraph{Replacement sampling in ELECTRA.}
Several works improve how ELECTRA's replacements are produced.
\citet{hao2021learning} observe that the jointly trained generator becomes
over-confident and add a hardness-prediction head that samples replacements
in proportion to the expected discriminator loss, together with a focal loss
for the generator. Fast-ELECTRA~\citep{dong2024fastelectra} removes the cost
of training the generator by using an existing language model as the
auxiliary model, and finds that its replacements are too hard unless its
output distribution is smoothed by a temperature that is decreased over
pre-training. We adopt the same two ideas, a fixed pre-trained generator and
a descending temperature schedule, and use a logarithmic schedule that spends
only a short phase on easy replacements.

\paragraph{Modern encoders.}
ModernBERT~\citep{warner2024modernbert} revisited the encoder recipe with
trillions of training tokens, 8192-token context, rotary embeddings,
alternating local/global attention and unpadding.
Ettin~\citep{weller2025ettin} trained paired encoder and decoder suites on
the same data, and mmBERT~\citep{marone2025mmbert} extended the ModernBERT
recipe to more than 1800 languages with an annealed language schedule. These
models use MLM. Our models instead use DeBERTa's disentangled relative
attention and an RTD objective, implemented with custom FlashAttention-style
kernels~\citep{dao2022flashattention,dao2023flashattention2}, which we
describe in \S\ref{sec:flash}.

\paragraph{RTD and embeddings.}
The weakness of RTD encoders on embedding tasks (\S\ref{sec:intro}) was
measured on BEIR~\citep{thakur2021beir} by~\citet{warner2024modernbert}.
\citet{rep2024electra} trace the poor STS performance of ELECTRA sentence
embeddings to the discriminator's last layer and repair it by
\emph{truncating} the model before fine-tuning. We address the problem
during pre-training instead: the full depth of the discriminator is kept and
its last layer is trained to preserve contextual token content.

\paragraph{Combining MLM and RTD.}
The complementary strengths of the two objectives have motivated attempts to
combine them, but the combination is not straightforward.
\citet{he2023debertav3} show that when the generator (trained with MLM) and
the discriminator (trained with RTD) share token embeddings, the two losses
pull the embeddings in different directions, which they call
\emph{tug-of-war} dynamics. Their remedy, gradient-disentangled embedding
sharing, stops the MLM gradient from reaching the discriminator's embeddings,
i.e.\ it resolves the conflict by separating the two objectives rather than
combining them. \citet{clark2020electra} study hybrids of MLM and RTD as
ablations, none of which surpasses pure RTD on GLUE. Conversely, after
observing DeBERTaV3's ``strong results \ldots\ on classification tasks but
weak ones on retrieval'', \citet{warner2024modernbert} name training that
leverages both MLM and RTD as a promising direction. \method is such a
combination, designed so that the two signals do not compete for the same
representation (\S\ref{sec:method:window}).

\section{Method}
\label{sec:method}

\begin{figure}[!htb]
\centering
\includegraphics[width=\linewidth]{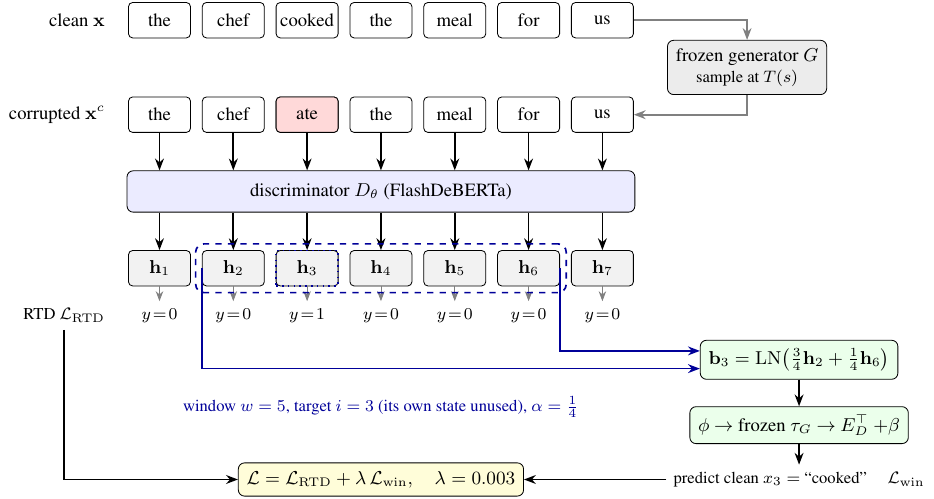}
\caption{One \method training step. The frozen generator replaces masked
positions by sampling at temperature $T(s)$. The discriminator reads the
corrupted sequence and is trained with token-level RTD on every position. In
addition, every token is reconstructed (as its \emph{clean} identity) from
only the two boundary states of a randomly sampled window that strictly
contains it, linearly interpolated by its relative offset. The two losses
are summed, with weight $\lambda=0.003$ on the window loss
(Eq.~\ref{eq:total}).}
\label{fig:window}
\end{figure}

\method trains a \emph{discriminator} encoder $D_\theta$ against a
\emph{frozen}, separately pre-trained MLM \emph{generator} $G$. Only the
discriminator encoder is released. Figure~\ref{fig:window} gives an overview
of one training step.

\subsection{Backbone}
\label{sec:method:backbone}

Both networks are DeBERTa-v2/v3 encoders~\citep{he2021deberta,he2023debertav3}
with disentangled relative attention (content-to-position and
position-to-content terms, shared key projections, 512 logarithmic position
buckets, LayerNorm on the relative embeddings) and \emph{no} absolute position
embeddings. All models use a factorized token embedding~\citep{lan2020albert}: a
128-dimensional token table followed by a linear projection to the hidden
size. With large
vocabularies, this keeps the embedding matrix small (16.4M parameters for the
English and 32.1M for the multilingual model) and leaves most of the
parameter budget for the Transformer layers. We reuse the tokenizers of
DeBERTaV3 (English) and mDeBERTa-v3 (multilingual). Table~\ref{tab:arch}
lists the configurations.

\begin{table}[!htb]
\centering
\small
\caption{Model configurations. Only the discriminator encoder is released;
the generator and the training heads are discarded after pre-training.}
\label{tab:arch}
\begin{tabular}{l cc cc}
\toprule
& \multicolumn{2}{c}{\textbf{\enmodel}} & \multicolumn{2}{c}{\textbf{\mlmodel}} \\
\cmidrule(lr){2-3}\cmidrule(lr){4-5}
& Discriminator & Generator & Discriminator & Generator \\
\midrule
Layers                     & 8     & 4     & 16    & 8 \\
Hidden / FFN / heads       & \multicolumn{4}{c}{384 / 1536 / 6} \\
Token embedding            & \multicolumn{4}{c}{factorized: 128-d table + linear projection to 384} \\
Relative attention         & \multicolumn{4}{c}{c2p + p2c, shared keys, 512 log buckets, LN on rel.\ embeddings} \\
Absolute positions         & \multicolumn{4}{c}{none} \\
Max.\ relative distance    & \multicolumn{4}{c}{2048 (longer distances share the outermost buckets)} \\
Dropout (hidden / attn.)   & \multicolumn{4}{c}{0.1 / 0.1} \\
Tokenizer                  & \multicolumn{2}{c}{DeBERTa-v3 (128,100)} & \multicolumn{2}{c}{mDeBERTa-v3 (251,000)} \\
\midrule
Parameters, total          & 31.0M & 24.2M & 61.0M & 46.8M \\
\quad non-embedding        & 14.2M & --    & 28.4M & -- \\
\midrule
RTD head (training only)   & 0.15M & & 0.15M & \\
Window head (training only)& 1.18M + 0.13M bias & & 1.18M + 0.25M bias & \\
\bottomrule
\end{tabular}
\end{table}

\paragraph{FlashDeBERTa.}
Both networks run on FlashDeBERTa (\S\ref{sec:flash}), our fused
implementation of disentangled attention. The attention formulation is
unchanged; FlashDeBERTa only avoids the dense $n\times n$ score matrices,
which is what makes 8192-token pre-training of a DeBERTa model practical.

\paragraph{Embedding initialisation.}
The discriminator encoder is initialised from scratch, except that its token
table and embedding projection are \emph{copied} from the pre-trained
generator. The two tables are not tied afterwards: the generator stays
frozen, and the discriminator's table is trained by both losses.

\subsection{Corruption with a frozen generator}
\label{sec:method:corruption}

Let $\mathbf{x}=(x_1,\dots,x_n)$ be a sequence of non-padding tokens. As in
BERT, a set $M$ of positions is selected with probability $p$ (special tokens
are never selected); 80\% of them are replaced by \texttt{[MASK]}, 10\% by a
random token and 10\% are left unchanged, giving $\tilde{\mathbf{x}}$. The
frozen generator predicts every selected position, and a replacement is
\emph{sampled} at temperature $T$:
\begin{equation}
  \hat{x}_i \sim \operatorname{softmax}\!\big(G(\tilde{\mathbf{x}})_i / T\big),
  \qquad
  x^{c}_i =
  \begin{cases}
    \hat{x}_i, & i\in M,\\
    x_i, & \text{otherwise.}
  \end{cases}
  \label{eq:corruption}
\end{equation}
A sample that equals the original token counts as original, so the RTD
target is $y_i=\mathbb{1}[x^c_i\neq x_i]$.

Unlike the original ELECTRA, where generator and discriminator are trained
jointly, we pre-train the generator first and freeze it, similarly to
Fast-ELECTRA~\citep{dong2024fastelectra}, which replaces the jointly trained
generator with an existing language model. This has three
practical advantages at our scale. (i) The difficulty of the replacements is
controlled explicitly by $T$ instead of by the generator's training dynamics.
(ii) The generator forward pass runs without gradients on a large
``superbatch'', so it costs a fraction of a discriminator step. (iii) The
generator's size can be chosen independently: we use half the
discriminator's depth (4 vs.\ 8 layers in English, 8 vs.\ 16 in the
multilingual model), within the 1/4--1/2 range recommended
by~\citet{clark2020electra}.

\paragraph{Hard-negative temperature curriculum.}
Early in training, the discriminator benefits from easy, noisy
replacements; later, it needs plausible ones. As in
Fast-ELECTRA~\citep{dong2024fastelectra} (\S\ref{sec:related}), we lower the
sampling temperature logarithmically during Phase~1
(\S\ref{sec:training:phase1}):
\begin{equation}
  T(s) = T_0 - c\,\ln\!\left(1+\frac{s}{s_0}\right), \qquad T_0=1.5,\quad c=0.1,
  \label{eq:temperature}
\end{equation}
where $s$ is the optimizer step and $s_0$ sets how fast the temperature
falls ($s_0\approx460$ steps for the English model and $\approx660$ for the
multilingual one). Most of the decrease happens in the first few percent of
training. For the English model, $T=1.39$ after 1k steps, $1.03$ after 50k
and $0.90$ at the end of Phase~1 (185k steps); for the multilingual model,
$T=1.41$, $1.07$ and $0.93$ at the end of Phase~1 (200k steps). Replacements
are thus increasingly drawn from the generator's high-probability region. In
Phase~2, the temperature stays at its final Phase-1 value: $0.9$ for the
English model and $0.93$ for the multilingual one.
To keep the curves comparable while $T$ changes,
all RTD diagnostics are computed on an independently corrupted copy of the
batch at a fixed $T=1$.

\subsection{Replaced-token detection}
\label{sec:method:rtd}

The discriminator reads the corrupted sequence,
$\mathbf{h}=D_\theta(\mathbf{x}^c)\in\mathbb{R}^{n\times d}$, and a small RTD head
(LayerNorm, Linear $d\to d$, GELU, Linear $d\to1$; 0.15M parameters)
predicts $y_i$ for every non-padding token:
\begin{equation}
  \mathcal{L}_{\mathrm{RTD}}
  = \frac{1}{n}\sum_{i=1}^{n}
    \mathrm{BCE}\big(\sigma(r(\mathbf{h}_i)),\,y_i\big).
  \label{eq:rtd}
\end{equation}

\subsection{Boundary-window reconstruction}
\label{sec:method:window}

$\mathcal{L}_{\mathrm{RTD}}$ alone gives the final layer no reason to keep the
identity of a token or the content of its neighbourhood
(\S\ref{sec:intro}). A token-level reconstruction objective such as MLM
would supply this information, but adding MLM directly to the discriminator
is problematic. Apart from the tug-of-war on shared
representations (\S\ref{sec:related}), the two objectives make conflicting
demands on the same position in a single forward pass: RTD needs the state
of a replaced token to signal that it is \emph{implausible}, whereas MLM
needs the same state to reconstruct the \emph{original} token, and MLM needs
\texttt{[MASK]} inputs that the RTD input does not contain.

The window task avoids both conflicts. A token is never reconstructed from
its own state, but \emph{only} from the final-layer states of two other
tokens, the boundaries of a window around it, and the task runs on the same
corrupted input in the same forward pass as RTD. Every token is a target and
also a boundary of the windows around its neighbours, so a final-layer state
only minimises the loss if it encodes the identities of the tokens around
it, in order. This is the content that a mean-pooled sentence embedding
aggregates. The target is always the \emph{clean} token, also at replaced
positions, so the boundaries must ``see through'' corruptions, which is
complementary to RTD's task of detecting them. The remaining tension between
the two objectives is controlled by a single weight $\lambda$
(Eq.~\ref{eq:total}).

\paragraph{Window sampling.}
For every position $i$ that is neither padding nor a special token, we sample a width $w$ uniformly from the
feasible widths in $[w_{\min},w_{\max}]=[3,12]$, and then a left boundary
$l$ uniformly such that $i$ lies \emph{strictly} inside the window:
\begin{equation}
  l \in \{\,i-w+2,\;\dots,\;i-1\,\}, \qquad r = l+w-1,
  \label{eq:window}
\end{equation}
where a width or offset is feasible only if the whole window
$[l,r]$ consists of real tokens of the same sequence. Boundaries may be
replaced tokens. The target $i$ itself is never visible to the
reconstruction, and its own hidden state is not used.

The width range was chosen empirically. Very short windows make the task
too easy: it is learned quickly and the window head even begins to overfit.
Very long windows make it too hard to provide a useful signal. A single fixed
width does not work either: reconstruction quality drops sharply as soon as
the distance between the target and the boundaries differs from the one seen
in training. We therefore sample the width (and the offset of the target
inside the window) for every token, so that each boundary state is trained to
encode its neighbours over the whole range of distances from 1 to
$w_{\max}-2=10$ tokens.

\paragraph{Boundary interpolation.}
The two boundary states are linearly interpolated by the relative offset of
the target inside the window,
\begin{equation}
  \alpha_i = \frac{i-l}{w-1}, \qquad
  \mathbf{b}_i = \operatorname{LN}\!\big((1-\alpha_i)\,\mathbf{h}_l
                 + \alpha_i\,\mathbf{h}_r\big),
  \label{eq:interp}
\end{equation}
so a target next to the left boundary is reconstructed mostly from
$\mathbf{h}_l$ and vice versa. No position embedding is added: the relative
position enters only through $\alpha_i$, so the boundary states themselves
must encode which tokens lie at which distance.

\paragraph{Decoding through the generator.}
$\mathbf{b}_i$ passes through a trainable FFN
$\phi(\mathbf{b})=W_2\,\mathrm{GELU}(W_1\mathbf{b}+c_1)+c_2$ with the backbone's
dimensions ($d\to4d\to d$), and is then decoded with the \emph{frozen}
generator's LM transform $\tau_G$ (a dense layer into the 128-dimensional
factorized embedding space, GELU, LayerNorm) against the
\emph{discriminator's} trainable token table $E_D\in\mathbb{R}^{|V|\times128}$ and
a trainable vocabulary bias $\beta$ (initialised from the generator's output
bias):
\begin{equation}
  p_\theta(\cdot \mid \mathbf{b}_i)
  = \operatorname{softmax}\!\Big(E_D\,\tau_G\big(\phi(\mathbf{b}_i)\big)+\beta\Big),
  \qquad
  \mathcal{L}_{\mathrm{win}}
  = -\frac{1}{|W|}\sum_{i\in W}\log p_\theta(x_i\mid\mathbf{b}_i),
  \label{eq:window-loss}
\end{equation}
where $W$ is the set of positions with a feasible window ($>99.5\%$ of tokens
in practice), and $x_i$ is the \emph{clean} original token, also at
positions that were replaced. Reusing the generator's transform keeps the
head tiny (1.18M parameters for $\phi$ and 0.13--0.25M for $\beta$), and
decoding against $E_D$ couples the auxiliary task to the input embeddings that
the released model keeps.

\paragraph{Total loss.}
\begin{equation}
  \mathcal{L} = \mathcal{L}_{\mathrm{RTD}} + \lambda\,\mathcal{L}_{\mathrm{win}},
  \qquad \lambda = 0.003.
  \label{eq:total}
\end{equation}
The weight $\lambda$ was chosen empirically, because the two objectives pull
in opposite directions. Increasing $\lambda$ weakens the RTD task and lowers
GLUE scores, but improves MTEB; decreasing it has the opposite effect. We
selected $\lambda=0.003$ as the operating point that keeps most of the GLUE
advantage of RTD while avoiding the poor embedding quality typical of pure
RTD discriminators. At this weight, RTD remains the dominant signal and the
window loss acts as a regulariser on the content of the final layer. The RTD head, $\phi$, the
window LayerNorm and $\beta$ are dropped after pre-training.

\section{Pre-training Data}
\label{sec:data}

We build two corpora with the same pipeline. The English corpus has
$\approx$1.0T tokens from fourteen sources, and the English model reads all
of it once. The multilingual corpus covers 46 languages and contains the
English corpus as its English portion; the multilingual model reads
$\approx$1.1T tokens of it.
Both are split into three disjoint, length-stratified stages for a
512$\to$2048$\to$8192 length curriculum.

\subsection{Three length stages}
\label{sec:data:stages}

\paragraph{Length-based routing.}
Documents are routed by an estimated token length
$\hat{\ell}=\lfloor|d|_{\mathrm{char}}/3\rfloor$:
\begin{equation}
  \mathrm{stage}(d)=
  \begin{cases}
    3 & \hat{\ell}>4096,\\
    2 & 1024<\hat{\ell}\le4096,\\
    1 & \hat{\ell}\le1024.
  \end{cases}
  \label{eq:routing}
\end{equation}
The ratio of 3 characters per token intentionally over-estimates length
(SentencePiece reaches 4--5 characters per token on typical English), so that
dense text such as code, mathematics or legal boilerplate is not routed into
a stage where it would be truncated. Character-based routing avoids a full
tokenization pass over the raw corpus solely for routing.

\paragraph{Overflow cascade.}
Each stage has a per-source token budget. When a stage's budget is exhausted,
further documents routed to it are split into non-overlapping chunks for the
next shorter stage (at most 2048 tokens for Stage~2, 512 for Stage~1); a chunk
is kept only if it is longer than half the target length,
$\ell_{\mathrm{chunk}}>L_{\max}/2$. Long documents therefore still contribute
to short-context training once the long-context budget is full. The stages
are disjoint at the document level. Sources with mostly short documents get
little or no Stage~3 budget, so Stage~3 is dominated by sources with naturally
long documents: legal text, full papers, PDFs and books.

\paragraph{Shuffling and storage.}
Within each source the order of raw shards is shuffled before tokenization,
so budgets are filled with a representative sample rather than a
chronological prefix. All shards of a stage are then merged and globally
shuffled with a two-pass sliding-window procedure (windows of 500 shards,
row-level shuffling within each window, a second pass over the already
shuffled output). Each row holds one document (or chunk), stored right-padded
to the stage length.

\subsection{English corpus}
\label{sec:data:en}

The corpus is designed to cover a broad range of registers, domains and
document lengths. Table~\ref{tab:data-en} lists the contribution of each
source to each stage.

\begin{table}[!htb]
\centering
\small
\setlength{\tabcolsep}{5pt}
\caption{Composition of the English corpus by source and stage (DeBERTa-v3 tokens,
budgeted). Percentages sum to 100 within each stage.}
\label{tab:data-en}
\begin{tabular}{ll rr rr rr}
\toprule
& & \multicolumn{2}{c}{\textbf{Stage 1} ($L=512$)}
& \multicolumn{2}{c}{\textbf{Stage 2} ($L=2048$)}
& \multicolumn{2}{c}{\textbf{Stage 3} ($L=8192$)} \\
\cmidrule(lr){3-4}\cmidrule(lr){5-6}\cmidrule(lr){7-8}
\textbf{Category} & \textbf{Source} & \% & B\,tok & \% & B\,tok & \% & B\,tok \\
\midrule
Web          & FineWeb-Edu      & 79.0 & 711.2 & 24.6 & 19.70 &  5.8 &  1.75 \\
News         & CC-News          &  0.8 &   7.0 &  3.8 &  3.00 &  0.0 &  0.00 \\
Sci.\ PDFs  & FinePDFs-Edu     &  4.1 &  36.5 & 14.4 & 11.50 & 20.0 &  6.00 \\
Reference    & Wikipedia EN     &  0.1 &   1.1 &  2.2 &  1.80 &  3.4 &  1.02 \\
Books        & Project Gutenberg&  0.2 &   2.2 &  2.5 &  2.00 &  6.7 &  2.00 \\
Legal        & Pile of Law      &  1.3 &  12.0 &  7.5 &  6.00 & 13.3 &  4.00 \\
Code         & StarCoderData    &  4.1 &  36.5 &  8.4 &  6.75 &  4.2 &  1.25 \\
Code         & SciCode          &  0.4 &   4.0 &  1.9 &  1.50 &  0.8 &  0.25 \\
Math         & AutoMathText     &  1.7 &  15.0 &  3.8 &  3.00 &  4.1 &  1.23 \\
Math         & Open-Web-Math    &  0.4 &   3.6 &  5.6 &  4.50 &  5.0 &  1.50 \\
Scientific   & peS2o            &  2.8 &  25.0 & 11.3 &  9.00 & 16.7 &  5.00 \\
Social/Q\&A  & Stack Exchange   &  2.6 &  23.7 &  1.9 &  1.50 &  0.8 &  0.25 \\
Scientific   & arXiv (TeX)      &  1.5 &  13.1 &  7.5 &  6.00 & 13.3 &  4.00 \\
Biomedical   & PubMed Central   &  1.0 &   9.1 &  4.7 &  3.75 &  6.7 &  2.00 \\
\midrule
\textbf{Total} & & \textbf{100} & \textbf{900.0} & \textbf{100} & \textbf{80.00} & \textbf{100} & \textbf{30.00} \\
\bottomrule
\end{tabular}
\end{table}

\paragraph{Web and news.}
FineWeb-Edu~\citep{penedo2024fineweb}, web text filtered from CommonCrawl by
a classifier trained on educational-quality annotations, is the dominant
source of Stage~1 (79\% of tokens) and is sharply reduced in later stages.
CC-News~\citep{hamborg2017news} adds contemporary factual prose with a high
density of named entities.

\paragraph{Long-form documents.}
FinePDFs-Edu~\citep{penedo2025finepdfs} provides text extracted from
educational and technical PDFs; because PDF text is dense and long, its share
grows from 4\% in Stage~1 to 20\% in Stage~3. English Wikipedia contributes
structured encyclopaedic prose at all stages. Project
Gutenberg\footnote{\url{https://www.gutenberg.org}} supplies public-domain books, which are chunked by
the overflow cascade (\S\ref{sec:data:stages}) and are most valuable at Stage~3. Pile of
Law~\citep{henderson2022pileoflaw} adds court opinions, contracts, statutes and
regulatory filings; contracts are particularly long (median $\approx$24K
characters), making this the most concentrated source of naturally long
documents.

\paragraph{Code and mathematics.}
StarCoderData~\citep{li2023starcoder} provides source code in more than 80
programming languages, and SciCode adds scientific code from open-access
repositories. Open-Web-Math~\citep{paster2023openwebmath} and
AutoMathText~\citep{zhang2024automathtext} cover mathematical web text and
self-contained mathematical exposition.

\paragraph{Scientific and biomedical text.}
peS2o~\citep{soldaini2023pes2o} (v3) contributes open-access papers and
abstracts from Semantic Scholar. arXiv preprints are included in their native
\LaTeX{} source, preserving notation and structural markup. PubMed Central
Open Access articles are parsed from JATS XML into a structured representation
that keeps the section hierarchy, abstract, figure captions and inline
citation markers and drops reference lists. Stack Exchange threads add
informal technical question--answer text.

\paragraph{Licences of scientific papers.}
We use only papers that are openly available for reuse. peS2o contains
open-access papers only, and from PubMed Central we take only the Open
Access subset (its commercial and non-commercial licence tiers), not the
full archive.

\subsection{Multilingual corpus}
\label{sec:data:ml}

The multilingual corpus combines the whole English corpus as its English
portion with three multilingual sources: FineWeb-2~\citep{penedo2025fineweb2}
for 45 non-English languages, and FinePDFs-Edu and the full
FinePDFs~\citep{penedo2025finepdfs} for 41 and 32 languages, respectively
(Table~\ref{tab:data-ml}). Web text dominates the short stage: FineWeb-2
provides half of the Stage~1 tokens and the two PDF sources only 2.5\%. In
the long stages the balance shifts towards PDFs, which reach 38\% of Stage~3,
because long documents in most languages come mainly from PDFs. All text is
tokenized with the mDeBERTa-v3 SentencePiece tokenizer.

\begin{table}[!htb]
\centering
\small
\setlength{\tabcolsep}{5pt}
\caption{Multilingual data read by the discriminator, by source and stage
(mDeBERTa-v3 tokens; expected values, since the training shards are shuffled).
Percentages sum to 100 within each stage. The per-language breakdown is in
Appendix~\ref{app:languages}.}
\label{tab:data-ml}
\begin{tabular}{lr rr rr rr}
\toprule
& & \multicolumn{2}{c}{\textbf{Stage 1} ($L=512$)}
& \multicolumn{2}{c}{\textbf{Stage 2} ($L=2048$)}
& \multicolumn{2}{c}{\textbf{Stage 3} ($L=8192$)} \\
\cmidrule(lr){3-4}\cmidrule(lr){5-6}\cmidrule(lr){7-8}
\textbf{Source} & \textbf{Lang.} & \% & B\,tok & \% & B\,tok & \% & B\,tok \\
\midrule
English corpus (14 sources) & 1 & 47.5 & 488.5 & 39.7 & 14.8 & 29.8 & 5.9 \\
FineWeb-2 & 45 & 50.0 & 513.6 & 44.9 & 16.7 & 31.9 & 6.3 \\
FinePDFs-Edu & 41 & 2.3 & 24.0 & 12.0 & 4.5 & 26.9 & 5.3 \\
FinePDFs (full) & 32 & 0.2 & 2.0 & 3.4 & 1.3 & 11.4 & 2.3 \\
\midrule
\textbf{Total} & 46 & \textbf{100} & \textbf{1028.1} & \textbf{100} & \textbf{37.2} & \textbf{100} & \textbf{19.8} \\
\bottomrule
\end{tabular}
\end{table}

\paragraph{Language selection.}
The model is aimed at English, the official languages of the European Union
and their neighbours (with particular attention to Slavic languages), and the
largest global markets: 46 languages in 10 scripts in total
(Appendix~\ref{app:languages}). The set includes all 24 official EU languages
and 12 Slavic languages. We also deliberately include Latin: it is no longer
spoken, but it remains the source of much scientific, medical and legal
vocabulary in European languages. Its text comes mostly from long PDF
documents, so its share is small in Stage~1 and grows in the long stages.

\paragraph{Language balance.}
English is kept as the largest single language (47.5\% of Stage~1) as the
main source of scientific and technical knowledge and of code; its share
falls to 39.7\% in Stage~2 and 29.8\% in Stage~3. The non-English budget is
distributed by temperature sampling, with each language's share proportional
to the square root of its available tokens, separately for web and PDF
sources. This reduces the gap between large and small languages: in Stage~1
the largest non-English language, Russian, has 4.7\% of the tokens and the
smallest, Yiddish, 0.05\%. FinePDFs-Edu is used first, and the full FinePDFs
only where the educational subset runs out.

\paragraph{Repetition and routing.}
Repetition is limited: documents may be repeated only in Stage~1, at most
five times for the smallest languages (Irish, Maltese, Swahili, Yiddish) and
at most twice for a few mid-sized ones; most languages are read only once,
and Stages~2 and~3 never repeat a document within a stage. Stage routing uses
a per-language characters-per-token estimate instead of the fixed English
ratio of \S\ref{sec:data:stages}, since this ratio differs widely between
scripts. The English shards are decoded and re-encoded with the mDeBERTa-v3
tokenizer.

\subsection{Resulting stages}
\label{sec:data:summary}

Table~\ref{tab:stages} summarises the three stages for both models. Padding
grows with length: 22.5\%, 27.5\% and 35.0\% of token slots in the English
Stages~1--3. For this reason, we report all budgets in \emph{real}
(non-padding) tokens.

\begin{table}[!htb]
\centering
\small
\setlength{\tabcolsep}{5pt}
\caption{The three length stages after tokenization. The English columns
describe the whole corpus, which the English model reads once. The
multilingual columns give the data read by the multilingual discriminator
(for Phase~2, the expected draw from the shuffled decay mix). ``Real tokens''
exclude padding; ``Mean len.'' is the mean number of real tokens per row.}
\label{tab:stages}
\begin{tabular}{l r rrr rrr}
\toprule
& & \multicolumn{3}{c}{\textbf{English}} & \multicolumn{3}{c}{\textbf{Multilingual}} \\
\cmidrule(lr){3-5}\cmidrule(lr){6-8}
Stage & $L_{\max}$ & Rows & Real tok. & Mean len. & Rows & Real tok. & Mean len. \\
\midrule
1 & 512   & 2{,}268M & 900.1B & 397   & 2{,}492M & 1{,}028.1B & 413 \\
2 & 2048  & 53.8M    & 79.8B  & 1{,}485 & 23.7M    & 37.2B      & 1{,}572 \\
3 & 8192  & 5.66M    & 30.2B  & 5{,}325 & 3.3M     & 19.8B      & 6{,}052 \\
\midrule
Total & & 2{,}328M & 1{,}010.1B & & 2{,}519M & 1{,}085.1B & \\
\bottomrule
\end{tabular}
\end{table}

\section{Training}
\label{sec:training}

Both models follow the same two-phase recipe. In \textbf{Phase~1}, the
generator and then the discriminator are pre-trained on the short-context
Stage~1 data (512 tokens), which accounts for roughly 90\% of the
discriminator's tokens. In \textbf{Phase~2}, the discriminator is continued on
the long-document Stages~2 and~3 (up to 8192 tokens) and its learning rate is
decayed. The two models differ only in how Phase~2 is organised
(\S\ref{sec:training:phase2}). Tables~\ref{tab:runs-en} and~\ref{tab:runs-ml}
list all runs.

\begin{table}[!htb]
\centering
\small
\setlength{\tabcolsep}{4pt}
\caption{Training runs of \enmodel. Each discriminator run continues from the
previous one; in Phase~2, Stage~2 and Stage~3 are trained one after the other
against the frozen generator with 512-token chunked attention. ``Real
tok./step'' excludes padding. WSD fractions are warmup / stable / decay in
\% of steps; the arrow gives the final learning rate relative to the peak.
$\times3$: learning-rate multiplier for the relative-position embeddings and
the RTD and window heads.}
\label{tab:runs-en}
\resizebox{\linewidth}{!}{%
\begin{tabular}{l l l r r r r l l r}
\toprule
Phase & Run & Data & $L$ & $p$ & Real tok./step & Steps & Peak LR & Schedule (WSD) & Real tokens \\
\midrule
\multirow{2}{*}{1} & generator     & Stage 1 & 512  & 0.3 & 4.88M & 18{,}500  & $6\times10^{-4}$ & 5/85/10 $\to0.1\times$ & 90.2B \\
 & discriminator & Stage 1 & 512  & 0.3 & 4.88M & 185{,}000 & $6\times10^{-4}$ & 5/85/10 $\to0.1\times$ & 902.0B \\
\midrule
\multirow{2}{*}{2} & discriminator & Stage 2 & 2048 & 0.3 & 4.56M & 17{,}500  & $6\times10^{-5}$, $\times3$ & 5/80/15 $\to0.5\times$ & 79.8B \\
 & discriminator & Stage 3 & 8192 & 0.3 & 4.09M & 7{,}370 & $3\times10^{-5}$, $\times3$ & 2/48/50 $\to0$ & 30.1B \\
\midrule
\multicolumn{9}{l}{Discriminator total} & 1{,}012B \\
\bottomrule
\end{tabular}}
\end{table}

\begin{table}[!htb]
\centering
\small
\setlength{\tabcolsep}{4pt}
\caption{Training runs of \mlmodel, in the format of Table~\ref{tab:runs-en}.
The Phase-1 discriminator has no decay; Phase~2 supplies it.
$^*$Decay mix: Stage-2 and Stage-3 rows plus Stage-1 rows not read in Phase~1
(20\% of the tokens).}
\label{tab:runs-ml}
\resizebox{\linewidth}{!}{%
\begin{tabular}{l l l r r r r l l r}
\toprule
Phase & Run & Data & $L$ & $p$ & Real tok./step & Steps & Peak LR & Schedule (WSD) & Real tokens \\
\midrule
\multirow{2}{*}{1} & generator     & Stage 1 & 512 & 0.3 & 5.07M & 39{,}490  & $6\times10^{-4}$ & 5/85/10 $\to0.1\times$ & 200.2B \\
 & discriminator & Stage 1 & 512 & 0.3 & 5.07M & 200{,}000 & $6\times10^{-4}$ & 7/93/0 & 1{,}013.9B \\
\midrule
2 & discriminator & decay mix\textsuperscript{*} & $\le$8192 & 0.3 & 7.12M & 10{,}000 & $6\times10^{-4}$ & 6/44/50 $\to0$ & 71.2B \\
\midrule
\multicolumn{9}{l}{Discriminator total} & 1{,}085B \\
\bottomrule
\end{tabular}}
\end{table}

\subsection{Common setup}
\label{sec:training:common}

All runs use PyTorch with Accelerate and FSDP2 on a single node with eight
GPUs: 8$\times$ NVIDIA RTX PRO 6000 Blackwell (96\,GB) for the English model
and 8$\times$ NVIDIA H100 (80\,GB) for the multilingual
model. The whole model is one FSDP unit (no auto-wrap, no CPU offload, no
resharding after the forward pass). We train in bf16 mixed precision with
TF32 matrix multiplications and fp32 master weights.

\paragraph{Unpadded computation.}
In all runs, generator and discriminator, the encoders run on real tokens
only (\S\ref{sec:flash}), and outputs and gradients match the padded
computation exactly. The padding in the stored stages
(\S\ref{sec:data:summary}) thus affects only the number of real tokens per
batch, not the compute spent per real token.

\paragraph{Optimisation.}
We use StableAdamW~\citep{wortsman2023stable} with
$\beta=(0.9,0.98)$, $\epsilon=10^{-6}$, decoupled weight decay 0.01 and no
gradient clipping, and a warmup--stable--decay (WSD)
schedule~\citep{hu2024minicpm,hagele2024scaling}: linear warmup, a constant
phase, and a $1-\sqrt{\cdot}$ decay,
\begin{equation}
  \eta(t)=\eta_{\max}\cdot
  \begin{cases}
    t/T_{\mathrm w}, & t<T_{\mathrm w},\\
    1, & T_{\mathrm w}\le t<T_{\mathrm w}+T_{\mathrm s},\\
    r+(1-r)\Big(1-\sqrt{(t-T_{\mathrm w}-T_{\mathrm s})/T_{\mathrm d}}\Big),
       & \text{otherwise,}
  \end{cases}
  \label{eq:wsd}
\end{equation}
where $r$ is the final-to-peak learning-rate ratio. Every run starts from a
fresh optimizer state.

\paragraph{Batch.}
Every Phase-1 run (generator and discriminator, both languages) processes
$8\ \text{GPUs}\times4\ \text{accumulation steps}\times384\ \text{rows}\times512
=6{,}291{,}456$ token slots per optimizer step, of which 4.88M (English) and
5.07M (multilingual) are real tokens. Since the vocabulary projection of a
250K-token vocabulary is the largest activation in the model, all
vocabulary projections (generator, window head) are computed in chunks of
8192 tokens, so logits never grow with the batch.

\subsection{Phase 1: short-context pre-training}
\label{sec:training:phase1}

The generator is trained from scratch with MLM at masking rate $p=0.3$ on
Stage-1 data (90B real tokens for the 4-layer English generator,
200B for the 8-layer multilingual one) and then frozen. The discriminator is
trained against the frozen generator at $p=0.3$, peak learning rate $6\times10^{-4}$ and the logarithmic
temperature schedule of Eq.~\eqref{eq:temperature}.

The English discriminator makes one full pass over Stage~1 (185{,}000 steps,
902B real tokens) and ends with a $1-\sqrt{\cdot}$ decay to $0.1\times$ the peak
rate over the last 10\% of steps. The multilingual discriminator is trained
for 200{,}000 steps (1.01T real tokens) at a
\emph{constant} learning rate after warmup, so that its only decay happens in
Phase~2, on long-context data.

\subsection{Phase 2: long-context training and decay}
\label{sec:training:phase2}

Phase~2 continues the discriminator on long documents and anneals it, at a
generator temperature of 0.9 (English) or 0.93 (multilingual;
\S\ref{sec:method:corruption}). The English model reads all of its
long-document data (Stages~2 and~3) once; the multilingual model reads 71.2B
tokens of a mix dominated by long documents. In both models, the generator is \emph{not} trained further: the
frozen Phase-1 generator is reused with chunked attention.

\paragraph{Chunked generator attention.}
The frozen generator is trained only on rows of at most 512 tokens, and its
predictions degrade sharply on much longer rows. Instead of continuing the
generator on long documents, we run it with block-diagonal attention over
independent 512-token segments in a single forward pass (cheaper than full
attention), while the discriminator attends over the full row. Corruption only needs plausible \emph{local} replacements, so a
512-token context is sufficient for the generator; in our preliminary
experiments, continuing the generator on long-context data made no noticeable
difference for the discriminator, and dropping it removes one training run per
long-context stage.

\paragraph{English model: two consecutive sub-stages.}
For the English model, Phase~2 runs over Stage~2 (2048 tokens, 79.8B real
tokens) and then Stage~3 (8192 tokens, 30.1B), each in one pass and without
Stage-1 replay, against the same frozen Phase-1 generator; the masking rate
stays at $p=0.3$. Since the Phase-1 checkpoint has already been decayed to
$6\times10^{-5}$, the Stage-2 run continues at that learning rate instead of
re-warming. The relative-position embeddings, whose far buckets were never
trained at 512 tokens, and the RTD and window heads receive a $3\times$
learning-rate multiplier. The Stage-3 run starts from the Stage-2 discriminator, continues at $3\times10^{-5}$ and decays to zero over its
second half.

\paragraph{Multilingual model: a single mixed decay run.}
The decay mix combines Stage~2 and Stage~3 with Stage-1 rows taken only from
shards that Phase~1 never reached, so no data is repeated. By tokens it is
20.0\% / 52.3\% / 27.8\% from Stages~1/2/3, with a mean row length of 1159,
stored \emph{without padding} and globally shuffled. Phase~2
starts from all Phase-1 weights with a fresh optimizer, re-warms over 580
steps to the peak learning rate, and decays it to zero over the last 5{,}000
of 10{,}000 steps (6{,}144 rows and $\approx$7.1M real tokens per step, 71.2B
tokens in total). The masking rate stays at $p=0.3$.

\section{Results}
\label{sec:results}

We evaluate the released discriminators on natural-language understanding
(GLUE~\citep{wang2019glue}), embedding tasks (MTEB English v2, MTEB Multilingual) and
zero-shot cross-lingual transfer (XTREME). Unless noted otherwise, results
are single runs with hyperparameters chosen on development data.

\subsection{GLUE}
\label{sec:results:glue}

\paragraph{Protocol.}
We follow the Ettin protocol~\citep{weller2025ettin}: learning rate
$\{1,3,5,8,10\}\times10^{-5}$, batch size $\{16,32\}$,
epochs $\{1,2,3,4\}$ for large tasks and $\{2,5,10,12\}$ for small ones, weight
decay $10^{-6}$, no warmup, maximum length 512, bf16. We report the best dev
score of the sweep. All runs use a single seed (42). MRPC, STS-B and RTE start from an
MNLI-adapted checkpoint: the model is fine-tuned on MNLI and interpolated back
into the pre-trained weights, as in WiSE-FT~\citep{wortsman2022robust} (MNLI share 0.9 for the body and word embeddings
and 1.0 for the relative-position buckets).

\begin{table}[!htb]
\centering
\small
\setlength{\tabcolsep}{4pt}
\caption{GLUE dev results. Our rows: best of the hyperparameter sweep with a single seed
(\S\ref{sec:results:glue}). Baseline rows are copied from the cited papers:
English baselines from Ettin~\citep{weller2025ettin} (same protocol as ours),
multilingual baselines from mmBERT~\citep{marone2025mmbert} (learning rate and
epochs swept, best per task), and DeBERTaV3-base from
ModernBERT~\citep{warner2024modernbert} (average only). CoLA: Matthews
correlation; STS-B: Spearman; MRPC: mean of accuracy and F1 for our models;
other tasks: accuracy. Params: total parameters.}
\label{tab:glue}
\resizebox{\linewidth}{!}{%
\begin{tabular}{l r rrrrrrrr r}
\toprule
Model & Params & CoLA & SST-2 & MRPC & STS-B & QQP & MNLI & QNLI & RTE & \textbf{Avg} \\
\midrule
\multicolumn{11}{l}{\textit{English encoders, $\le$35M parameters}} \\
BERT-small~\citep{turc2019wellread}  & 28.8M & 44.8 & 90.1 & 83.1 & 87.6 & 90.1 & 79.2 & 88.2 & 69.3 & 79.0 \\
MiniLM-L12~\citep{wang2020minilm}    & 33.4M & 59.1 & 93.3 & 91.2 & 89.2 & 91.5 & 85.6 & 91.9 & 74.7 & 84.6 \\
Ettin-Enc-32m~\citep{weller2025ettin}& 31.9M & 57.4 & 92.0 & 89.7 & 89.5 & 91.0 & 83.4 & 90.7 & 74.7 & 83.5 \\
\textbf{\enmodel}                    & 31.0M & 64.9 & 92.2 & 91.8 & 90.2 & 91.1 & 84.5 & 90.3 & 82.7 & \textbf{86.0} \\
\midrule
\multicolumn{11}{l}{\textit{Larger English encoders (reference)}} \\
DistilBERT~\citep{sanh2019distilbert}& 66.4M & 56.9 & 92.2 & 86.8 & 87.4 & 90.8 & 82.7 & 89.5 & 66.1 & 81.5 \\
Ettin-Enc-68m~\citep{weller2025ettin}& 68.1M & 64.8 & 94.4 & 92.2 & 91.1 & 91.9 & 87.0 & 92.9 & 83.8 & 87.2 \\
BERT-base~\citep{devlin2019bert}     & 109.5M & 59.0 & 93.1 & 89.5 & 89.4 & 91.4 & 85.4 & 91.6 & 78.2 & 84.7 \\
DeBERTaV3-base~\citep{he2023debertav3} & 184M & \multicolumn{8}{c}{} & 88.1 \\
ModernBERT-base~\citep{warner2024modernbert} & 149.0M & 65.1 & 96.0 & 92.2 & 91.8 & 92.1 & 89.1 & 93.9 & 87.4 & 88.4 \\
\midrule
\multicolumn{11}{l}{\textit{Multilingual encoders}} \\
mDistilBERT~\citep{sanh2019distilbert}         & 134.7M & 34.7 & 89.4 & 90.5 & 87.5 & 86.6 & 79.0 & 87.5 & 73.3 & 77.5 \\
Multilingual MiniLM~\citep{wang2020minilm}     & 117.7M & 25.4 & 91.6 & 91.3 & 88.6 & 87.7 & 82.1 & 89.7 & 75.8 & 78.3 \\
mmBERT-small~\citep{marone2025mmbert}          & 140.5M & 61.8 & 93.1 & 91.6 & 90.3 & 88.6 & 85.8 & 91.9 & 81.9 & 84.7 \\
\textbf{\mlmodel}                              & 61.0M  & 58.3 & 91.4 & 92.1 & 90.2 & 91.5 & 84.8 & 91.2 & 81.6 & \textbf{85.1} \\
\bottomrule
\end{tabular}}

\end{table}

\paragraph{Results.}
Among English encoders of comparable size (Table~\ref{tab:glue}), \enmodel
reaches 86.0 GLUE, ahead of Ettin-Enc-32m (83.5), evaluated with the same
protocol, MiniLM-L12 (84.6) and BERT-small (79.0).
It is also ahead of BERT-base (84.7) and DistilBERT (81.5), and 1.2 points
behind Ettin-Enc-68m (87.2), which has twice its parameters. The largest gains
are on CoLA (+7.5 over Ettin-Enc-32m) and RTE (+8.0), where RTD encoders are
known to be strong. \mlmodel reaches 85.1, ahead of the 2.3$\times$ larger
mmBERT-small (84.7) and far ahead of Multilingual MiniLM (78.3) and mDistilBERT
(77.5). Despite spending more than half of its pre-training on other
languages, it is only 0.9 points behind our English model, with the gap
concentrated in CoLA ($-6.6$).

\subsection{Embedding tasks: MTEB}
\label{sec:results:mteb}

\paragraph{Protocol.}
Following mmBERT~\citep{marone2025mmbert}, each backbone is fine-tuned once
into a sentence embedding model on MS\,MARCO~\citep{nguyen2016msmarco}
(query, positive, mined hard negative) triplets\footnote{\url{https://huggingface.co/datasets/sentence-transformers/msmarco-co-condenser-margin-mse-sym-mnrl-mean-v1},
subset \texttt{triplet-hard}.} with Sentence-Transformers~\citep{reimers2019sbert}. Training is
contrastive: we use the multiple-negatives ranking
(InfoNCE) loss~\citep{henderson2017efficient,oord2018cpc} with
in-batch negatives plus the hard negative of each triplet, scale 20, computed
with gradient caching~\citep{gao2021gradcache} over an effective batch of
1024 triplets. We train for one epoch on 1M triplets with mean pooling, 5\%
warmup, bf16, and learning rate $5\times10^{-4}$ (English model) or
$3\times10^{-4}$ (multilingual model). We report the main metric of each task, in percent, on the test split, averaged over
the subsets of each task, for MTEB English v2 (41
tasks)~\citep{muennighoff2023mteb,enevoldsen2025mmteb} and MTEB Multilingual (131
tasks)~\citep{enevoldsen2025mmteb}.

\begin{table}[!htb]
\centering
\small
\setlength{\tabcolsep}{4pt}
\caption{MTEB English v2 (41 tasks), main metric of each task in percent, per category.
Every backbone is fine-tuned once on MS\,MARCO before evaluation. Baseline
rows are copied from Ettin~\citep{weller2025ettin} and
mmBERT~\citep{marone2025mmbert}; our fine-tuning follows the mmBERT protocol.
Mean (Type) is the mean over the seven categories, Mean (Task) over the 41
tasks (not reported by mmBERT).}
\label{tab:mteb-en}
\resizebox{\linewidth}{!}{%
\begin{tabular}{l r rrrrrrr rr}
\toprule
Model & Params & Class. & Clust. & PairCl. & Rerank. & Retr. & STS & Summ. & \textbf{Mean (Type)} & Mean (Task) \\
\midrule
\multicolumn{11}{l}{\textit{English encoders, $\le$35M parameters}} \\
BERT-small~\citep{turc2019wellread}   & 28.8M & 64.7 & 39.6 & 79.4 & 41.7 & 38.1 & 73.0 & 21.6 & 51.1 & 54.0 \\
MiniLM-L12~\citep{wang2020minilm}     & 33.4M & 64.9 & 37.8 & 79.8 & 43.5 & 38.4 & 73.2 & 21.4 & 51.3 & 53.9 \\
Ettin-Enc-32m~\citep{weller2025ettin} & 31.9M & 64.2 & 39.6 & 77.1 & 42.6 & 39.7 & 73.2 & 19.9 & 50.9 & 54.2 \\
\textbf{\enmodel}                     & 31.0M & 65.3 & 39.7 & 78.0 & 42.3 & 38.6 & 73.1 & 20.9 & \textbf{51.1} & 54.2 \\
\midrule
\multicolumn{11}{l}{\textit{Larger English encoders (reference)}} \\
Ettin-Enc-68m~\citep{weller2025ettin} & 68.1M & 66.6 & 40.1 & 79.3 & 43.3 & 43.1 & 74.2 & 21.7 & 52.6 & 56.1 \\
BERT-base~\citep{devlin2019bert}      & 109.5M & 67.2 & 40.4 & 80.5 & 43.1 & 41.2 & 74.8 & 23.1 & 52.9 & 56.0 \\
ModernBERT-base~\citep{warner2024modernbert} & 149.0M & 67.5 & 41.3 & 80.4 & 44.7 & 43.9 & 75.3 & 25.2 & 54.0 & 57.1 \\
\midrule
\multicolumn{11}{l}{\textit{Multilingual encoders}} \\
Multilingual MiniLM~\citep{wang2020minilm} & 117.7M & 59.3 & 36.9 & 77.3 & 42.4 & 35.0 & 71.9 & 19.3 & 48.9 & -- \\
mDistilBERT~\citep{sanh2019distilbert}     & 134.7M & 59.0 & 38.3 & 75.9 & 42.7 & 38.4 & 72.4 & 19.4 & 49.4 & -- \\
mmBERT-small~\citep{marone2025mmbert}      & 140.5M & 62.1 & 40.8 & 78.9 & 44.1 & 40.7 & 74.1 & 23.6 & 52.1 & -- \\
\textbf{\mlmodel}                          & 61.0M  & 63.8 & 38.6 & 76.8 & 43.0 & 37.9 & 72.1 & 20.6 & \textbf{50.4} & 53.3 \\
\bottomrule
\end{tabular}}
\end{table}

\paragraph{English.}
After a single MS\,MARCO pass, \enmodel reaches 51.1 MTEB English v2
(Table~\ref{tab:mteb-en}), on par with the MLM encoders of its size:
Ettin-Enc-32m (50.9), BERT-small (51.1) and MiniLM-L12 (51.3). Its retrieval
score (38.6 nDCG@10) is in the same range (38.1--39.7). This is the category
where RTD backbones are known to fail under the same kind of fine-tuning
(\S\ref{sec:intro}). The \method discriminator does not show this collapse:
its embeddings match MLM encoders of the same size while it is clearly
stronger on GLUE. \mlmodel reaches 50.4 on English MTEB, ahead of
Multilingual MiniLM (48.9) and mDistilBERT (49.4) and 1.7 points behind
mmBERT-small (52.1), which is 2.3$\times$ larger.

\begin{table}[!htb]
\centering
\small
\setlength{\tabcolsep}{4pt}
\caption{MTEB Multilingual v2 (131 tasks), main metric of each task in percent, per
category, after one MS\,MARCO fine-tuning epoch. Baseline rows are copied
from mmBERT~\citep{marone2025mmbert}. Avg is the mean over the eight
categories shown, as in mmBERT; our model additionally scores $-3.3$ on the
three instruction-reranking tasks (mean over all nine categories: 38.5; mean
over tasks: 44.5). MTEB Multilingual covers far more languages than the 46 of
our pre-training corpus.}
\label{tab:mteb-ml}
\resizebox{\linewidth}{!}{%
\begin{tabular}{l r rrrrrrrr r}
\toprule
Model & Params & Bitext & PairCl. & Class. & STS & Retr. & MultiLab. & Clust. & Rerank. & \textbf{Avg} \\
\midrule
mDistilBERT~\citep{sanh2019distilbert}     & 134.7M & 37.2 & 76.6 & 50.3 & 62.1 & 36.7 & 14.6 & 37.5 & 62.2 & 47.1 \\
Multilingual MiniLM~\citep{wang2020minilm} & 117.7M & 46.6 & 76.7 & 51.0 & 64.6 & 35.6 & 14.0 & 34.9 & 64.1 & 48.4 \\
mmBERT-small~\citep{marone2025mmbert}      & 140.5M & 50.5 & 77.4 & 50.4 & 64.8 & 41.9 & 15.5 & 38.7 & 66.1 & 50.7 \\
\textbf{\mlmodel}                          & 61.0M  & 25.0 & 71.9 & 48.3 & 58.5 & 41.3 & 15.3 & 35.4 & 54.5 & 43.8 \\
\bottomrule
\end{tabular}}
\end{table}

\paragraph{Multilingual.}
On MTEB Multilingual (Table~\ref{tab:mteb-ml}), \mlmodel is competitive on
retrieval (41.3, vs.\ 41.9 for mmBERT-small and 35.6--36.7 for Multilingual
MiniLM and mDistilBERT) and multilabel classification, but behind all three
baselines on average (43.8 vs.\ 47.1--50.7). The gap comes mostly from bitext
mining (25.0 vs.\ 37.2--50.5) and reranking (54.5 vs.\ 62.2--66.1). Both
categories include many languages outside our 46-language corpus, which
the baselines (100+ and 1800+ languages) cover; bitext mining in addition
depends on cross-lingual alignment that the English-only MS\,MARCO
fine-tuning does not provide.

\subsection{Zero-shot cross-lingual transfer: XTREME}
\label{sec:results:xtreme}

\paragraph{Protocol.}
Models are fine-tuned on English and evaluated on all languages: XNLI (trained
on MNLI), PAWS-X, XCOPA (trained on SIQA, then COPA), XQuAD, MLQA and
TyDiQA-GoldP (one SQuAD v1.1 model), WikiANN and UDPOS
(\citealp{hu2020xtreme}). We compare with Multilingual
MiniLM~\citep{wang2020minilm}, mmBERT-small~\citep{marone2025mmbert} and
multilingual DistilBERT~\citep{sanh2019distilbert}, all run through the same
pipeline with identical hyperparameters: learning rate
$\{1,2,3,5,8\}\times10^{-5}$, batch size 32 (16 for XCOPA), maximum length 128
(384 for QA), warmup 10\%, no weight decay, bf16; 3 epochs for XNLI and QA, 5
for PAWS-X, 10 for WikiANN and UDPOS, 3 on SIQA followed by 10 on COPA at
$10^{-5}$ for XCOPA. The learning rate and the checkpoint are selected on the
\emph{English} dev set only; test scores are never used for selection. We
use a single seed (42), so differences of about one point are within noise,
in particular on XCOPA (500 test examples per language). MLQA is evaluated on
same-language question--context pairs, and tagging labels the first subword
of each word.

\begin{table}[!htb]
\centering
\small
\setlength{\tabcolsep}{4pt}
\caption{XTREME zero-shot cross-lingual transfer (fine-tuned on English).
XNLI, PAWS-X, XCOPA: accuracy; XQuAD, MLQA, TyDiQA-GoldP: F1; WikiANN: entity
F1; UDPOS: token accuracy. Top: the 26 XTREME languages covered by our
pre-training corpus. Bottom: all languages. Single seed; learning rate and
checkpoint selected on English dev data.}
\label{tab:xtreme}
\resizebox{\linewidth}{!}{%
\begin{tabular}{l r rrrrrrrr r}
\toprule
Model & Params & XNLI & PAWS-X & XCOPA & XQuAD & MLQA & TyDiQA & WikiANN & UDPOS & \textbf{Avg} \\
\midrule
\multicolumn{11}{l}{\textit{Languages covered by our pre-training (26)}} \\
\mlmodel              & 61.0M  & 71.5 & 83.5 & 59.1 & 69.4 & 62.0 & 62.0 & 57.4 & 77.9 & \textbf{67.8} \\
Multilingual MiniLM   & 117.7M & 71.1 & 83.2 & 58.8 & 67.2 & 61.1 & 58.5 & 58.9 & 79.4 & 67.3 \\
mmBERT-small          & 140.5M & 73.8 & 85.4 & 62.8 & 73.0 & 66.2 & 66.8 & 59.4 & 72.4 & 70.0 \\
mDistilBERT           & 134.7M & 60.8 & 79.6 & 53.5 & 50.1 & 45.5 & 40.3 & 59.7 & 72.5 & 57.7 \\
\midrule
\multicolumn{11}{l}{\textit{All XTREME languages}} \\
\mlmodel              & 61.0M  & 69.0 & 83.5 & 55.6 & 64.7 & 62.0 & 57.3 & 45.5 & 64.3 & 62.7 \\
Multilingual MiniLM   & 117.7M & 70.9 & 83.2 & 58.3 & 66.3 & 61.1 & 58.3 & 55.2 & 77.8 & 66.4 \\
mmBERT-small          & 140.5M & 73.6 & 85.4 & 60.3 & 71.6 & 66.2 & 66.3 & 52.0 & 69.5 & 68.1 \\
mDistilBERT           & 134.7M & 59.3 & 79.6 & 52.7 & 47.4 & 45.5 & 39.3 & 53.9 & 68.8 & 55.8 \\
\bottomrule
\end{tabular}}
\end{table}

\paragraph{Results.}
\mlmodel covers 46 languages, 26 of which appear in XTREME. On these
languages (Table~\ref{tab:xtreme}, top), it averages 67.8, ahead of
Multilingual MiniLM (67.3) and mDistilBERT (57.7), and 2.2 points behind
mmBERT-small (70.0), despite being 1.9--2.3$\times$ smaller than all three
baselines. It is ahead of Multilingual MiniLM on six of the eight tasks (all
but WikiANN and UDPOS) and ahead of mmBERT-small on UDPOS. Over all XTREME
languages (bottom), the average drops to 62.7, because the model has never
seen 16 of them (Thai, Tamil, Telugu, Malayalam, Burmese, Georgian, Kazakh,
Marathi, Basque, Persian, Afrikaans, Tagalog, Javanese, Yoruba, Haitian and
Quechua), several in scripts that are absent from its pre-training corpus;
on token-level tasks in these languages (WikiANN, UDPOS) it falls far behind
models pre-trained on $\approx$100 languages or more. Selecting the learning
rate on the test set instead gives the same ranking (covered languages: 68.7
vs.\ 68.1 MiniLM, 70.8 mmBERT-small, 58.4 mDistilBERT). Per-language results
are in Appendix~\ref{app:xtreme}.

\section{FlashDeBERTa}
\label{sec:flash}

All models in this paper are trained and evaluated with FlashDeBERTa, our implementation of
DeBERTa's disentangled attention as fused GPU kernels. The library has been
open source since 2025, and this paper is its first description. It loads existing DeBERTa-v2 and DeBERTa-v3 checkpoints
without retraining and reproduces the outputs of the reference
implementation.

\paragraph{Why standard kernels do not apply.}
In DeBERTa, the attention score of a query token $i$ and a key token $j$ has
three terms:
\begin{equation}
  s_{ij} = \mathbf{q}_i^\top\mathbf{k}_j
         + C_{i,\,\sigma(i,j)} + P_{j,\,\sigma(i,j)},
  \qquad
  C = Q R_k^\top,\quad P = K R_q^\top .
  \label{eq:disentangled}
\end{equation}
Here $R_k$ and $R_q$ are the projected relative-position embeddings, with one
row per position slot, and $\sigma(i,j)$ is the slot of the bucketed relative
distance between $i$ and $j$. $C$ is the content-to-position term and $P$ the
position-to-content term. The reference implementation gathers $C$ and $P$
into two dense $n\times n$ matrices and adds them to a dense $n\times n$
content matrix, for every head. Fused kernels such as
FlashAttention~\citep{dao2022flashattention,dao2023flashattention2} avoid the
$n\times n$ matrix, but they accept only queries, keys and values, so the two
position terms cannot be passed to them.

\paragraph{Forward kernel.}
FlashDeBERTa computes $C$ and $P$ as two ordinary matrix products. Each has
one row per token and one column per position slot, so their size grows
linearly with the sequence length. A Triton
kernel~\citep{tillet2019triton} then processes the attention in tiles of
queries and keys. For each tile it computes the content scores, reads the
matching entries of $C$ and $P$, and updates a running softmax and the
weighted sum of values, as in FlashAttention. The slot $\sigma(i,j)$ is read
from a small table indexed by the distance, so the logarithmic bucket is not
recomputed for every pair. No $n\times n$ tensor is stored at any point.

\begin{figure}[!tb]
\centering
\includegraphics[width=\linewidth]{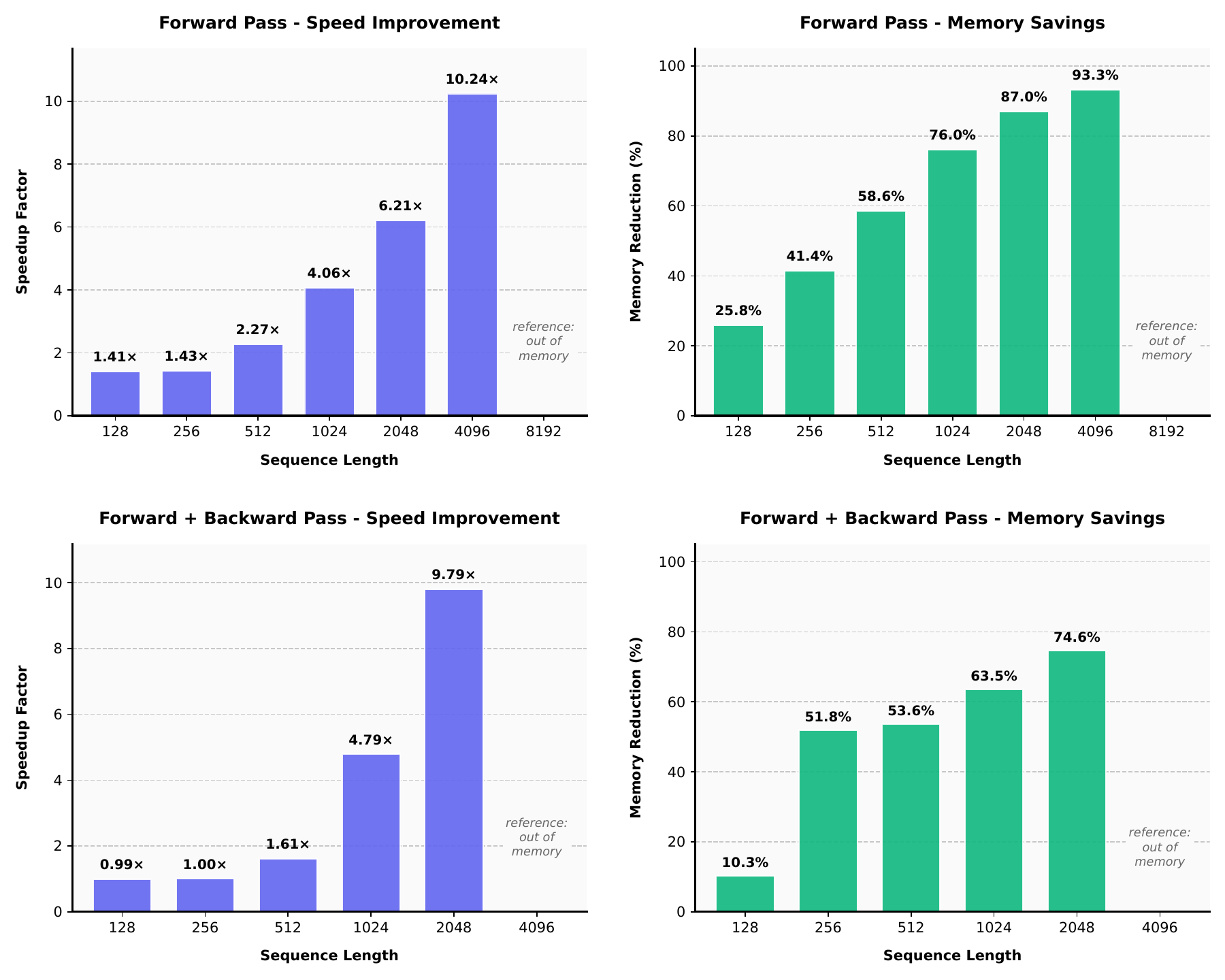}
\caption{FlashDeBERTa against the reference implementation in
\texttt{transformers} on DeBERTa-v3-base at batch size 4 in BF16, on an
NVIDIA L4 GPU: speed-up (left) and reduction of activation memory (right)
for the forward pass (top) and for the forward and backward pass (bottom).
At the last length of each row the reference implementation runs out of
memory.}
\label{fig:flash-gliclass}
\end{figure}

\paragraph{Backward kernels.}
Two further kernels compute the gradients of keys and values and of queries.
They recompute the scores tile by tile, so the attention matrix is not kept
from the forward pass either. The gradients of the position scores are
accumulated per distance: every (token, distance) cell has a single writer,
so no atomic operations are needed, and a one-hot matrix then maps distances
back to position slots, which turns the gradients of $R_k$ and $R_q$ into a
few batched matrix products. The kernel structure is derived from
FlagAttention~\citep{flagattention}, and the distance lookup and the banded
gradient accumulation are adapted from
DisentangledFlash~\citep{disentangledflash}.

\paragraph{Batches without padding.}
When a batch contains padding, the encoder removes it once and keeps the real
tokens of all sequences end to end. Projections, LayerNorms, feed-forward
blocks and attention then run only on real tokens, and the padded shape is
restored at the output. The attention kernel takes the sequence boundaries,
and the mapping from kernel tiles to sequences is built on the GPU, so no
synchronisation with the host is needed. On a half-padded batch this is about
twice as fast.

\paragraph{Other features.}
The kernels are registered as custom operators, so the models run under
\texttt{torch.compile} and CUDA graphs in both inference and training. The
attention products can optionally run in FP8. Token positions can be given
explicitly, which allows layouts other than a sequence, such as a 2-D grid of
image patches. Tile sizes that do not fit the shared memory of a GPU are
reduced automatically.

\paragraph{Benchmarks.}
Figure~\ref{fig:flash-gliclass} compares FlashDeBERTa with the reference
implementation on DeBERTa-v3-base at batch size 4 in BF16. In the forward
pass the speed-up grows from 1.4$\times$ at 128 tokens to 2.3$\times$ at 512
and 10.2$\times$ at 4096, and the memory saving from 26\% to 59\% and 93\%.
The reference implementation runs out of memory at 8192 tokens, while
FlashDeBERTa processes 8192 tokens in 0.87\,s with 1.2\,GB and 16384 tokens
in 2.9\,s with 2.4\,GB. For the forward and backward pass together, the two
implementations are equally fast up to 256 tokens; the speed-up is
1.6$\times$ at 512 tokens and 9.8$\times$ at 2048, with 54\% and 75\% less
memory. The reference implementation runs out of memory at 4096 tokens, and
FlashDeBERTa trains at 8192 tokens with 12\,GB.

\paragraph{Availability.}
FlashDeBERTa is available under the Apache~2.0 licence at
\url{https://github.com/Knowledgator/FlashDeBERTa} and as the
\texttt{flashdeberta} package. It provides the base model and heads for
masked language modelling, sequence and token classification, question
answering and multiple choice.

\section{Discussion and Limitations}
\label{sec:discussion}

\paragraph{No controlled ablation.}
While choosing the window-loss weight we observed a consistent trade-off
between GLUE and MTEB (\S\ref{sec:method:window}), but we report no
systematic ablation: training a discriminator without the window loss on the
same data was beyond our budget. Our evidence that \method avoids the RTD
embedding gap is therefore indirect: the documented failure of RTD backbones
on retrieval and STS
\citep{warner2024modernbert,rep2024electra,breton2025neobert,antoun2025modernbert}
against the MTEB results of our two models. A small-scale controlled
ablation (window weight, width range, interpolation vs.\ concatenation of
the boundaries) is the most important follow-up.

\paragraph{Evaluation.}
MTEB follows the backbone protocol of ModernBERT, Ettin and mmBERT: a single
contrastive fine-tuning pass on MS\,MARCO. The scores therefore measure the
backbone, not a fully trained embedding model. XTREME uses a single seed.
The GLUE and MTEB baseline numbers are copied from the Ettin and mmBERT
papers, whose protocols we follow, but learning-rate selection and minor
details differ, so differences of a few tenths of a point should not be
over-interpreted.

\paragraph{Language coverage.}
\mlmodel targets 46 languages. Languages outside this set, and in particular
scripts absent from pre-training (e.g.\ Thai, Tamil, Telugu, Malayalam,
Burmese, Georgian), are not supported, as the all-language XTREME results
show.

\paragraph{Phase-2 variants.}
The two models organise Phase~2 differently (\S\ref{sec:training:phase2}).
The single mixed decay run of the multilingual model is simpler and needs one
fewer run, but our evaluation does not isolate its effect from that of model
size and data.

\section{Conclusion}
\label{sec:conclusion}

We presented \method, an ELECTRA-style pre-training recipe that keeps
replaced-token detection as its main objective and adds boundary-window
reconstruction, a lightweight auxiliary task that makes every final-layer
state encode its neighbourhood. Together with a frozen, separately trained
generator, a hard-negative temperature curriculum, FlashDeBERTa kernels and a
512$\to$8192 length curriculum, it produces compact encoders that are strong
on classification, embedding and cross-lingual tasks: the English \enmodel
and the 46-language \mlmodel. We release both
models.\footnote{\url{https://huggingface.co/knowledgator/EdgeCoder-31M} and
\url{https://huggingface.co/knowledgator/EdgeCoder-Multilingual-61M}.} This
report also serves as the reference for FlashDeBERTa, the open-source fused
implementation of disentangled attention used to train them
(\url{https://github.com/Knowledgator/FlashDeBERTa}).

\section*{Acknowledgements}
The computing resources for training the models in this work were provided
through a grant from Google for Startups. We thank Tymon Kokoszka of Google
for Startups for his support throughout the programme. Our pre-training
framework grew out of earlier experiments that were run on the NVIDIA Brev
platform with credits provided by NVIDIA. We thank Stefan Schweter for
valuable discussions and feedback, which helped us choose the direction of
this work.

\bibliographystyle{plainnat}
\bibliography{references}

\clearpage
\appendix
\section{Pre-training hyperparameters}
\label{app:hparams}

\begin{table}[H]
\centering
\small
\caption{Hyperparameters shared by all discriminator runs unless stated
otherwise in Tables~\ref{tab:runs-en} and~\ref{tab:runs-ml}. The generator
temperature schedule is given in \S\ref{sec:method:corruption}
(Eq.~\ref{eq:temperature}).}
\label{tab:hparams}
\begin{tabular}{ll}
\toprule
\textbf{Component} & \textbf{Value} \\
\midrule
Objective & $\mathcal{L}_{\mathrm{RTD}} + 0.003\,\mathcal{L}_{\mathrm{win}}$ (RTD weight 1.0) \\
Masking & $p=0.3$ in all runs (generators and discriminators); 80/10/10 \\
Window widths & $w\in[3,12]$, uniform over feasible widths; target strictly inside \\
Window decoding & LN $\to$ FFN ($384\to1536\to384$) $\to$ frozen generator transform $\to$ $E_D$ + bias \\
Logit chunk size & 8192 targets, activation checkpointing \\
Diagnostic temperature & 1.0 (independent corruption) \\
Embedding init.\ & token table and projection copied from the frozen generator \\
\midrule
Optimizer & StableAdamW, $\beta=(0.9,0.98)$, $\epsilon=10^{-6}$, weight decay 0.01 \\
Gradient clipping & none \\
Schedule & WSD with $1-\sqrt{\cdot}$ decay (Eq.~\ref{eq:wsd}) \\
Peak LR (Phase 1) & $6\times10^{-4}$ \\
Batch (Phase 1) & $8\times4\times384$ rows of 512 tokens = 6.29M token slots \\
Precision & bf16 autocast, TF32 matmuls, fp32 master weights \\
Parallelism & FSDP2, single unit, no offload, no reshard after forward \\
Data loading & 16 shards prefetched per rank, rows shuffled across them; exact resume \\
Seeds & 42 (discriminators), 43 (multilingual generator) \\
\bottomrule
\end{tabular}
\end{table}

\section{Language coverage of \mlmodel}
\label{app:languages}

Table~\ref{tab:languages} lists the 46 pre-training languages with their
share of Stage~1 and of the decay mix, and the expected number of
tokens of each language read by the multilingual discriminator in Phase~1 and
Phase~2 (share $\times$ real tokens per step $\times$ steps). Languages
are sorted by their share of the decay mix.

\begingroup
\footnotesize
\setlength{\tabcolsep}{4pt}
\begin{longtable}{r l l l rr rr}
\caption{Pre-training languages of \mlmodel (mDeBERTa-v3 tokens, B $=10^9$).}
\label{tab:languages} \\
\toprule
& & & & \multicolumn{2}{c}{\textbf{Stage 1 / Phase 1}} & \multicolumn{2}{c}{\textbf{Decay mix / Phase 2}} \\
\cmidrule(lr){5-6}\cmidrule(lr){7-8}
\# & ISO & Language & Script & Share \% & Disc.\ B & Share \% & Disc.\ B \\
\midrule
\endfirsthead
\toprule
\# & ISO & Language & Script & Share \% & Disc.\ B & Share \% & Disc.\ B \\
\midrule
\endhead
\bottomrule
\endfoot
1 & \texttt{eng} & English & Latin & 47.51 & 481.8 & 38.50 & 27.41 \\
2 & \texttt{rus} & Russian & Cyrillic & 4.68 & 47.5 & 4.99 & 3.56 \\
3 & \texttt{cmn} & Chinese (Mandarin) & Han & 3.76 & 38.1 & 3.72 & 2.65 \\
4 & \texttt{spa} & Spanish & Latin & 2.96 & 30.1 & 3.70 & 2.63 \\
5 & \texttt{deu} & German & Latin & 3.05 & 31.0 & 3.65 & 2.60 \\
6 & \texttt{fra} & French & Latin & 2.79 & 28.3 & 3.49 & 2.48 \\
7 & \texttt{jpn} & Japanese & Japanese & 2.55 & 25.8 & 2.97 & 2.12 \\
8 & \texttt{ita} & Italian & Latin & 2.24 & 22.7 & 2.72 & 1.94 \\
9 & \texttt{por} & Portuguese & Latin & 1.99 & 20.1 & 2.52 & 1.79 \\
10 & \texttt{pol} & Polish & Latin & 1.72 & 17.5 & 2.04 & 1.46 \\
11 & \texttt{nld} & Dutch & Latin & 1.57 & 15.9 & 1.88 & 1.34 \\
12 & \texttt{vie} & Vietnamese & Latin & 1.59 & 16.1 & 1.66 & 1.18 \\
13 & \texttt{ind} & Indonesian & Latin & 1.41 & 14.3 & 1.56 & 1.11 \\
14 & \texttt{arb} & Arabic & Arabic & 1.21 & 12.3 & 1.49 & 1.06 \\
15 & \texttt{ces} & Czech & Latin & 1.20 & 12.2 & 1.48 & 1.05 \\
16 & \texttt{tur} & Turkish & Latin & 1.32 & 13.4 & 1.46 & 1.04 \\
17 & \texttt{hun} & Hungarian & Latin & 1.16 & 11.7 & 1.43 & 1.02 \\
18 & \texttt{swe} & Swedish & Latin & 1.15 & 11.7 & 1.37 & 0.98 \\
19 & \texttt{ron} & Romanian & Latin & 1.13 & 11.5 & 1.35 & 0.96 \\
20 & \texttt{ukr} & Ukrainian & Cyrillic & 1.04 & 10.5 & 1.30 & 0.92 \\
21 & \texttt{kor} & Korean & Hangul & 1.15 & 11.6 & 1.27 & 0.90 \\
22 & \texttt{ell} & Greek & Greek & 1.00 & 10.2 & 1.17 & 0.84 \\
23 & \texttt{nob} & Norwegian Bokmål & Latin & 1.04 & 10.6 & 1.16 & 0.82 \\
24 & \texttt{dan} & Danish & Latin & 0.99 & 10.0 & 1.13 & 0.80 \\
25 & \texttt{fin} & Finnish & Latin & 0.97 & 9.8 & 1.12 & 0.80 \\
26 & \texttt{slk} & Slovak & Latin & 0.78 & 7.9 & 0.95 & 0.67 \\
27 & \texttt{bul} & Bulgarian & Cyrillic & 0.77 & 7.8 & 0.89 & 0.63 \\
28 & \texttt{hin} & Hindi & Devanagari & 0.66 & 6.7 & 0.78 & 0.56 \\
29 & \texttt{hrv} & Croatian & Latin & 0.55 & 5.6 & 0.74 & 0.53 \\
30 & \texttt{bos} & Bosnian & Latin & 0.61 & 6.2 & 0.73 & 0.52 \\
31 & \texttt{lit} & Lithuanian & Latin & 0.62 & 6.3 & 0.72 & 0.52 \\
32 & \texttt{slv} & Slovenian & Latin & 0.56 & 5.6 & 0.70 & 0.50 \\
33 & \texttt{heb} & Hebrew & Hebrew & 0.60 & 6.1 & 0.68 & 0.49 \\
34 & \texttt{ben} & Bengali & Bengali & 0.55 & 5.6 & 0.62 & 0.44 \\
35 & \texttt{srp} & Serbian & Cyrillic & 0.39 & 3.9 & 0.58 & 0.41 \\
36 & \texttt{lvs} & Latvian & Latin & 0.48 & 4.9 & 0.56 & 0.40 \\
37 & \texttt{ekk} & Estonian & Latin & 0.52 & 5.2 & 0.55 & 0.39 \\
38 & \texttt{zsm} & Malay & Latin & 0.44 & 4.5 & 0.52 & 0.37 \\
39 & \texttt{urd} & Urdu & Arabic & 0.32 & 3.3 & 0.42 & 0.30 \\
40 & \texttt{mkd} & Macedonian & Cyrillic & 0.25 & 2.5 & 0.30 & 0.21 \\
41 & \texttt{lat} & Latin & Latin & 0.05 & 0.5 & 0.28 & 0.20 \\
42 & \texttt{bel} & Belarusian & Cyrillic & 0.23 & 2.3 & 0.26 & 0.19 \\
43 & \texttt{mlt} & Maltese & Latin & 0.12 & 1.2 & 0.18 & 0.13 \\
44 & \texttt{swh} & Swahili & Latin & 0.15 & 1.5 & 0.18 & 0.13 \\
45 & \texttt{gle} & Irish & Latin & 0.13 & 1.3 & 0.16 & 0.12 \\
46 & \texttt{ydd} & Yiddish & Hebrew & 0.05 & 0.5 & 0.07 & 0.05 \\
\midrule
 &  & \textbf{Total} &  & 100 & 1{,}013.9 & 100 & 71.2 \\
\end{longtable}
\endgroup

\section{XTREME per-language results}
\label{app:xtreme}

Per-language zero-shot transfer results. $^\dagger$: language not in our
pre-training corpus. QA tasks are given as F1\,/\,EM.

\begin{table}[H]
\centering
\footnotesize
\caption{XTREME per-language results: XNLI.}
\label{tab:xtreme-xnli}
\begin{tabular}{l rrrr}
\toprule
Lang. & \mlmodel & Multiling.\ MiniLM & mmBERT-small & mDistilBERT \\
\midrule
ar & 67.3 & 68.8 & 71.7 & 59.3 \\
bg & 74.8 & 74.2 & 75.9 & 62.3 \\
de & 74.4 & 73.0 & 77.4 & 65.8 \\
el & 72.7 & 73.0 & 73.1 & 58.1 \\
en & 84.7 & 81.5 & 85.7 & 79.0 \\
es & 77.1 & 75.3 & 79.4 & 69.7 \\
fr & 76.8 & 74.2 & 79.1 & 68.9 \\
hi & 65.2 & 66.5 & 69.0 & 50.0 \\
ru & 74.2 & 72.0 & 75.0 & 63.1 \\
sw & 60.1 & 63.4 & 62.7 & 45.3 \\
th$^\dagger$ & 33.9 & 67.8 & 70.7 & 37.9 \\
tr & 70.1 & 68.5 & 72.3 & 53.8 \\
ur & 61.7 & 63.4 & 62.6 & 52.6 \\
vi & 69.6 & 71.6 & 75.8 & 58.4 \\
zh & 72.6 & 69.8 & 73.9 & 64.8 \\
\bottomrule
\end{tabular}
\end{table}

\begin{table}[H]
\centering
\footnotesize
\caption{XTREME per-language results: PAWS-X.}
\label{tab:xtreme-pawsx}
\begin{tabular}{l rrrr}
\toprule
Lang. & \mlmodel & Multiling.\ MiniLM & mmBERT-small & mDistilBERT \\
\midrule
de & 87.8 & 85.8 & 88.8 & 82.3 \\
en & 94.8 & 94.1 & 95.1 & 91.6 \\
es & 87.5 & 87.4 & 88.5 & 83.3 \\
fr & 89.6 & 87.4 & 90.3 & 83.3 \\
ja & 75.6 & 75.5 & 79.2 & 72.5 \\
ko & 70.1 & 73.1 & 73.9 & 68.2 \\
zh & 78.7 & 79.5 & 81.6 & 76.0 \\
\bottomrule
\end{tabular}
\end{table}

\begin{table}[H]
\centering
\footnotesize
\caption{XTREME per-language results: XCOPA.}
\label{tab:xtreme-xcopa}
\begin{tabular}{l rrrr}
\toprule
Lang. & \mlmodel & Multiling.\ MiniLM & mmBERT-small & mDistilBERT \\
\midrule
et & 55.2 & 56.8 & 57.0 & 51.6 \\
ht$^\dagger$ & 52.4 & 54.8 & 52.4 & 47.4 \\
id & 62.4 & 57.8 & 67.2 & 54.2 \\
it & 57.0 & 59.0 & 68.2 & 53.6 \\
qu$^\dagger$ & 47.8 & 53.4 & 52.6 & 49.4 \\
sw & 56.2 & 56.4 & 52.6 & 52.6 \\
ta$^\dagger$ & 50.4 & 60.4 & 56.2 & 52.4 \\
th$^\dagger$ & 47.4 & 60.8 & 62.4 & 56.2 \\
tr & 57.4 & 59.4 & 59.8 & 50.4 \\
vi & 60.0 & 59.0 & 64.6 & 53.8 \\
zh & 65.6 & 63.2 & 70.2 & 58.0 \\
\bottomrule
\end{tabular}
\end{table}

\begin{table}[H]
\centering
\footnotesize
\caption{XTREME per-language results: XQuAD (F1 / EM).}
\label{tab:xtreme-xquad}
\begin{tabular}{l rrrr}
\toprule
Lang. & \mlmodel & Multiling.\ MiniLM & mmBERT-small & mDistilBERT \\
\midrule
ar & 62.4\,/\,45.9 & 61.9\,/\,43.5 & 67.1\,/\,49.7 & 44.0\,/\,28.8 \\
de & 74.2\,/\,58.1 & 71.3\,/\,56.1 & 77.5\,/\,61.2 & 55.6\,/\,39.3 \\
el & 68.2\,/\,51.8 & 64.3\,/\,45.0 & 68.4\,/\,51.7 & 40.8\,/\,24.8 \\
en & 84.6\,/\,73.6 & 80.7\,/\,69.5 & 85.6\,/\,74.4 & 78.9\,/\,66.7 \\
es & 75.1\,/\,57.9 & 71.2\,/\,52.7 & 80.1\,/\,61.7 & 61.0\,/\,41.3 \\
hi & 61.6\,/\,45.7 & 61.6\,/\,45.2 & 67.3\,/\,51.3 & 34.5\,/\,20.8 \\
ru & 74.0\,/\,57.0 & 71.5\,/\,54.8 & 74.2\,/\,56.5 & 59.2\,/\,42.0 \\
th$^\dagger$ & 17.9\,/\,8.7 & 57.5\,/\,43.7 & 57.4\,/\,44.0 & 20.3\,/\,14.6 \\
tr & 63.2\,/\,46.7 & 60.0\,/\,43.6 & 67.9\,/\,51.1 & 36.2\,/\,20.0 \\
vi & 66.8\,/\,46.7 & 67.9\,/\,49.6 & 76.2\,/\,57.3 & 47.0\,/\,28.4 \\
zh & 63.7\,/\,55.8 & 61.1\,/\,51.8 & 65.5\,/\,54.4 & 43.6\,/\,30.9 \\
\bottomrule
\end{tabular}
\end{table}

\begin{table}[H]
\centering
\footnotesize
\caption{XTREME per-language results: MLQA (F1 / EM).}
\label{tab:xtreme-mlqa}
\begin{tabular}{l rrrr}
\toprule
Lang. & \mlmodel & Multiling.\ MiniLM & mmBERT-small & mDistilBERT \\
\midrule
ar & 51.1\,/\,33.3 & 52.4\,/\,32.6 & 56.9\,/\,37.5 & 34.1\,/\,18.3 \\
de & 60.9\,/\,45.7 & 61.0\,/\,46.2 & 65.1\,/\,50.2 & 45.9\,/\,31.6 \\
en & 80.7\,/\,67.9 & 79.1\,/\,66.1 & 81.6\,/\,68.2 & 75.7\,/\,62.6 \\
es & 66.9\,/\,49.3 & 62.8\,/\,44.5 & 70.6\,/\,52.2 & 51.0\,/\,32.9 \\
hi & 53.9\,/\,37.6 & 55.1\,/\,38.5 & 58.1\,/\,40.8 & 28.3\,/\,14.8 \\
vi & 58.7\,/\,39.1 & 60.4\,/\,40.3 & 66.0\,/\,45.0 & 37.6\,/\,20.1 \\
zh & 61.7\,/\,39.7 & 57.1\,/\,34.9 & 65.1\,/\,40.5 & 46.1\,/\,25.3 \\
\bottomrule
\end{tabular}
\end{table}

\begin{table}[H]
\centering
\footnotesize
\caption{XTREME per-language results: TyDiQA (F1 / EM).}
\label{tab:xtreme-tydiqa}
\begin{tabular}{l rrrr}
\toprule
Lang. & \mlmodel & Multiling.\ MiniLM & mmBERT-small & mDistilBERT \\
\midrule
ar & 59.4\,/\,42.6 & 61.1\,/\,43.2 & 66.5\,/\,49.3 & 38.4\,/\,22.5 \\
bn & 57.7\,/\,42.5 & 39.1\,/\,23.9 & 53.3\,/\,38.9 & 31.3\,/\,18.6 \\
en & 71.7\,/\,63.0 & 70.8\,/\,60.2 & 75.4\,/\,66.8 & 68.5\,/\,58.9 \\
fi & 63.6\,/\,50.1 & 59.3\,/\,46.7 & 69.2\,/\,57.2 & 33.2\,/\,19.6 \\
id & 69.1\,/\,54.9 & 58.1\,/\,41.4 & 70.7\,/\,56.1 & 39.1\,/\,23.9 \\
ko & 50.0\,/\,39.1 & 53.2\,/\,40.6 & 60.8\,/\,49.6 & 18.3\,/\,9.8 \\
ru & 64.2\,/\,45.6 & 62.6\,/\,43.3 & 68.8\,/\,47.4 & 56.4\,/\,40.8 \\
sw & 60.1\,/\,47.7 & 63.7\,/\,54.3 & 69.5\,/\,56.1 & 37.3\,/\,24.4 \\
te$^\dagger$ & 20.2\,/\,13.3 & 57.2\,/\,43.6 & 62.8\,/\,45.3 & 30.8\,/\,23.9 \\
\bottomrule
\end{tabular}
\end{table}

\begin{table}[H]
\centering
\footnotesize
\caption{XTREME per-language results: WikiANN.}
\label{tab:xtreme-wikiann}
\begin{tabular}{l rrrr}
\toprule
Lang. & \mlmodel & Multiling.\ MiniLM & mmBERT-small & mDistilBERT \\
\midrule
af$^\dagger$ & 64.4 & 71.9 & 69.1 & 72.5 \\
ar & 38.1 & 37.7 & 34.7 & 36.8 \\
bg & 70.5 & 72.7 & 69.3 & 73.0 \\
bn & 51.6 & 61.6 & 55.4 & 54.2 \\
de & 70.5 & 69.8 & 75.4 & 75.1 \\
el & 61.6 & 65.7 & 58.2 & 59.4 \\
en & 82.1 & 81.6 & 82.7 & 82.8 \\
es & 72.8 & 65.2 & 80.1 & 65.7 \\
et & 66.4 & 69.4 & 64.8 & 70.4 \\
eu$^\dagger$ & 30.5 & 52.5 & 44.7 & 54.2 \\
fa$^\dagger$ & 18.6 & 41.4 & 38.1 & 33.1 \\
fi & 69.9 & 72.0 & 71.9 & 73.0 \\
fr & 73.6 & 74.8 & 76.0 & 75.6 \\
he & 34.0 & 43.4 & 37.8 & 47.3 \\
hi & 48.0 & 58.8 & 56.1 & 53.9 \\
hu & 67.0 & 70.6 & 69.7 & 69.3 \\
id & 50.3 & 47.9 & 49.0 & 51.2 \\
it & 73.5 & 74.7 & 76.3 & 78.4 \\
ja & 23.7 & 20.4 & 20.5 & 23.6 \\
jv$^\dagger$ & 39.7 & 47.4 & 50.9 & 54.5 \\
ka$^\dagger$ & 12.0 & 56.3 & 38.2 & 54.5 \\
kk$^\dagger$ & 22.5 & 41.5 & 35.0 & 42.3 \\
ko & 31.9 & 38.2 & 41.3 & 28.0 \\
ml$^\dagger$ & 0.9 & 51.4 & 30.5 & 32.6 \\
mr$^\dagger$ & 22.4 & 47.5 & 37.1 & 42.6 \\
ms & 51.6 & 51.8 & 68.0 & 66.5 \\
my$^\dagger$ & 0.0 & 54.7 & 30.1 & 31.5 \\
nl & 76.4 & 77.0 & 78.3 & 78.3 \\
pt & 73.2 & 73.1 & 78.3 & 74.7 \\
ru & 56.1 & 57.8 & 55.5 & 56.9 \\
sw & 62.9 & 57.2 & 62.5 & 71.8 \\
ta$^\dagger$ & 3.5 & 42.7 & 31.8 & 30.8 \\
te$^\dagger$ & 2.6 & 43.8 & 23.0 & 32.9 \\
th$^\dagger$ & 7.8 & 5.9 & 4.8 & 0.0 \\
tl$^\dagger$ & 60.7 & 69.7 & 69.0 & 74.2 \\
tr & 67.7 & 68.2 & 68.1 & 64.8 \\
ur & 28.5 & 30.6 & 27.2 & 26.6 \\
vi & 61.6 & 64.2 & 64.9 & 56.9 \\
yo$^\dagger$ & 42.9 & 49.8 & 32.5 & 48.3 \\
zh & 29.5 & 25.7 & 23.6 & 36.8 \\
\bottomrule
\end{tabular}
\end{table}

\begin{table}[H]
\centering
\footnotesize
\caption{XTREME per-language results: UDPOS.}
\label{tab:xtreme-udpos}
\begin{tabular}{l rrrr}
\toprule
Lang. & \mlmodel & Multiling.\ MiniLM & mmBERT-small & mDistilBERT \\
\midrule
af$^\dagger$ & 73.5 & 88.9 & 84.0 & 85.9 \\
ar & 64.8 & 69.6 & 66.4 & 58.7 \\
bg & 89.5 & 89.5 & 74.1 & 85.5 \\
de & 88.4 & 89.5 & 86.6 & 86.3 \\
el & 59.8 & 67.2 & 50.0 & 33.2 \\
en & 96.2 & 95.9 & 96.4 & 95.7 \\
es & 88.0 & 89.2 & 85.1 & 88.1 \\
et & 84.6 & 88.2 & 72.4 & 75.9 \\
eu$^\dagger$ & 28.0 & 77.8 & 57.7 & 67.9 \\
fa$^\dagger$ & 47.3 & 71.7 & 72.4 & 70.4 \\
fi & 84.3 & 87.2 & 73.8 & 76.9 \\
fr & 78.0 & 79.2 & 72.3 & 76.2 \\
he & 77.0 & 81.6 & 71.7 & 77.5 \\
hi & 75.1 & 72.0 & 61.9 & 64.1 \\
hu & 85.5 & 85.0 & 74.6 & 78.6 \\
id & 77.2 & 76.8 & 75.9 & 75.8 \\
it & 90.0 & 90.1 & 86.0 & 89.3 \\
ja & 53.6 & 54.3 & 45.0 & 43.4 \\
kk$^\dagger$ & 34.1 & 82.5 & 56.8 & 73.3 \\
ko & 63.0 & 65.4 & 59.8 & 53.8 \\
mr$^\dagger$ & 46.5 & 85.6 & 70.6 & 61.5 \\
nl & 89.1 & 90.0 & 88.9 & 88.9 \\
pt & 88.2 & 89.5 & 87.3 & 88.6 \\
ru & 84.8 & 85.5 & 77.1 & 80.5 \\
ta$^\dagger$ & 17.5 & 68.7 & 55.6 & 51.4 \\
te$^\dagger$ & 29.7 & 86.4 & 57.7 & 61.3 \\
th$^\dagger$ & 6.0 & 68.1 & 61.2 & 25.9 \\
tl$^\dagger$ & 27.7 & 89.4 & 87.3 & 72.9 \\
tr & 77.1 & 78.9 & 72.6 & 68.3 \\
ur & 70.3 & 67.8 & 54.5 & 61.5 \\
vi & 60.2 & 64.6 & 64.9 & 56.7 \\
yo$^\dagger$ & 19.6 & 21.4 & 25.5 & 32.6 \\
zh & 66.7 & 70.0 & 68.0 & 63.2 \\
\bottomrule
\end{tabular}
\end{table}

\end{document}